\documentclass[letterpaper, 10 pt, conference]{ieeeconf}  %

\usepackage[T1]{fontenc} %

\IEEEoverridecommandlockouts                              %

\usepackage{graphicx}
\usepackage{epsfig} %
\usepackage{mathptmx} %
\usepackage{times} %
\usepackage{amsmath} %
\usepackage{amssymb}  %
\usepackage{float}
\usepackage{capt-of}
\usepackage{booktabs}
\usepackage{multicol}
\usepackage{multirow}
\usepackage{pifont}
\usepackage{url}
\usepackage{hyperref}
\newcommand{\cmark}{\ding{51}}
\newcommand{\xmark}{\ding{55}}
\newcommand{\modelname}[1]{Nav-R1}
\newcommand{\datasetname}[1]{Nav-CoT-110K}
\title{\LARGE \bf
\raisebox{-0.5em}{\includegraphics[height=1.8em]{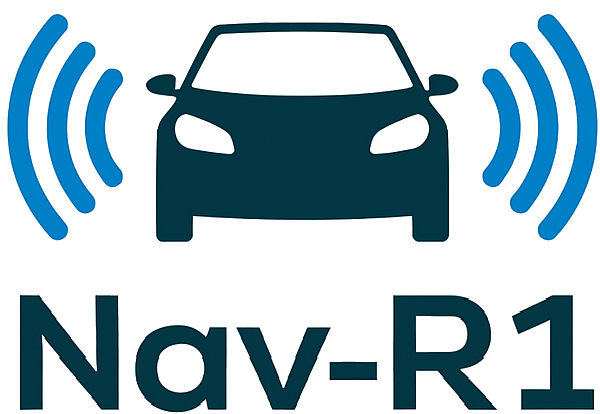}}~\modelname{}: Reasoning and Navigation in Embodied Scenes
}

\author{
    \textbf{Qingxiang Liu}$^{1*}$\quad 
    \textbf{Ting Huang}$^{1*}$\quad 
    \textbf{Zeyu Zhang}$^{2*\dag}$\quad 
    \textbf{Hao Tang}$^{2\ddag}$ \vspace{0.1cm}\\
    $^1$Shanghai University of Engineering Science\quad
    $^2$Peking University \vspace{0.05cm}\\
    \small $^*$Equal contribution. $^\dag$Project lead.
    $^\ddag$Corresponding author: bjdxtanghao@gmail.com.
}

\begin{document}

\makeatletter
\let\@oldmaketitle\@maketitle%
\renewcommand{\@maketitle}{\@oldmaketitle%
\vspace{0.2cm}
\begin{center}
    \textit{“The division of labor between System 1 (fast) and System 2 (slow) is highly efficient: it minimizes effort and optimizes performance.” --- Daniel Kahneman (Nobel Prize in Economics)}
\end{center}
{\centering
\includegraphics[width=\linewidth]{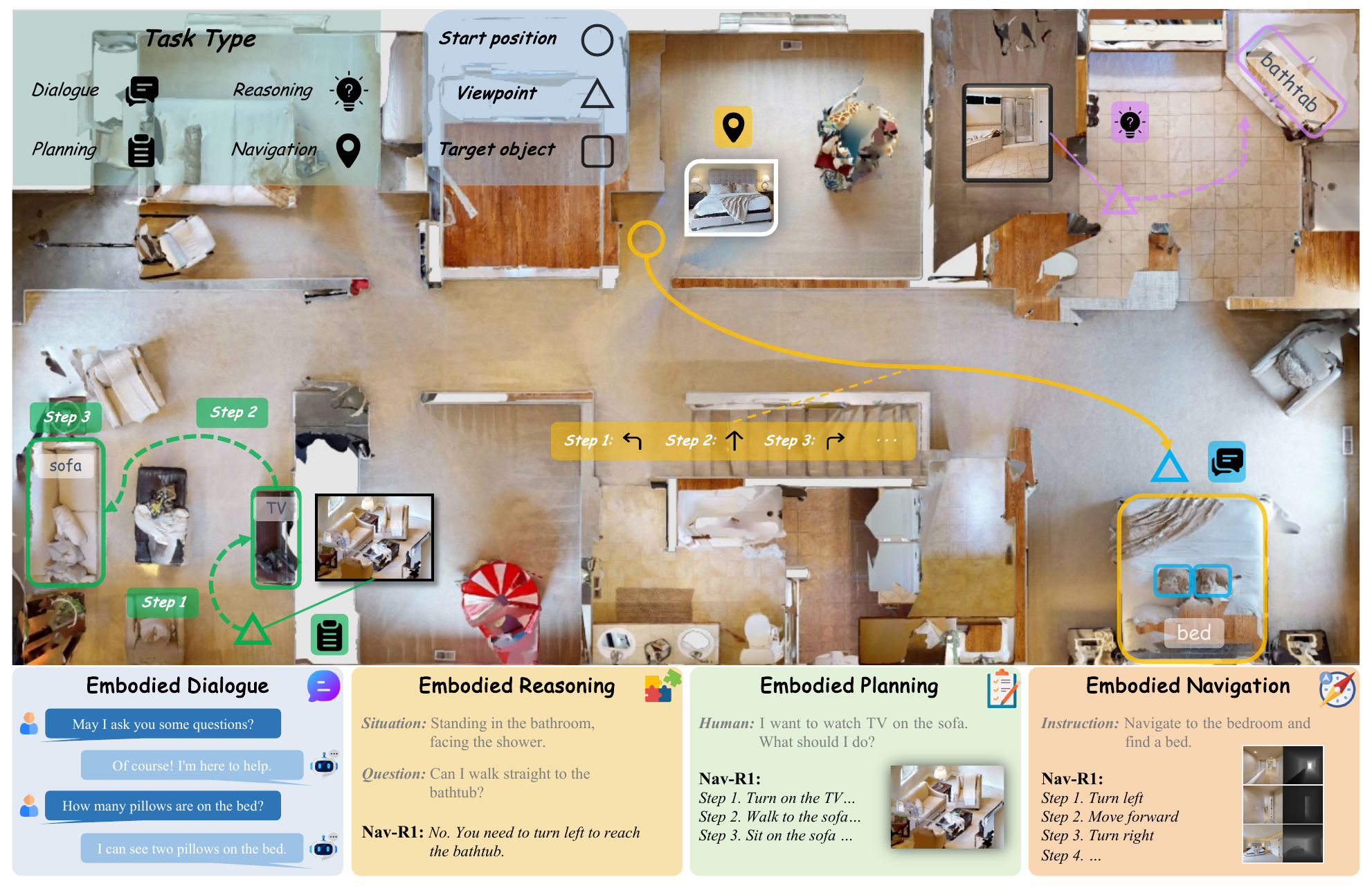}\par
\captionof{figure}{\textbf{\modelname{}} is an embodied foundation model that integrates dialogue, reasoning, planning, and navigation capabilities to enable intelligent interaction and task execution in 3D environments.}
\label{fig:main}}
\vspace{-0.5cm}
\bigskip}%
\makeatother

\maketitle
\setcounter{figure}{1}
\thispagestyle{empty}
\pagestyle{empty}

\begin{abstract}
Embodied navigation requires agents to integrate perception, reasoning, and action for robust interaction in complex 3D environments.
Existing approaches often suffer from incoherent and unstable reasoning traces that hinder generalization across diverse environments, and difficulty balancing long-horizon semantic reasoning with low-latency control for real-time navigation.
To address these challenges, we propose \modelname{}, an embodied foundation model that unifies reasoning in embodied environments.
We first construct \datasetname{}, a large-scale dataset of step-by-step Chains-of-Thought (CoT) for embodied tasks, which enables cold-start initialization with structured reasoning.
Building on this foundation, we design a GRPO-based reinforcement learning framework with three complementary rewards: format, understanding, and navigation, to improve structural adherence, semantic grounding, and path fidelity.
Furthermore, we introduce a Fast-in-Slow reasoning paradigm, decoupling deliberate semantic reasoning from low-latency reactive control for efficient yet coherent navigation.
Extensive evaluations on embodied AI benchmarks demonstrate that \modelname{} consistently outperforms strong baselines, with over 8\% average improvement in reasoning and navigation performance.
Real-world deployment on a mobile robot further validates its robustness under limited onboard resources.
Code: \url{https://github.com/AIGeeksGroup/Nav-R1}.
Website: \url{https://aigeeksgroup.github.io/Nav-R1}.
\end{abstract}

\section{Introduction}
Embodied scene understanding is a central problem in embodied AI, robotics, and intelligent agents, requiring an agent to perceive, reason, and act within complex 3D environments~\cite{anderson2018vision,krantz2020beyond}.
A robust understanding of embodied navigation not only supports goal-directed tasks such as object search, instruction following, and trajectory planning, but also enables higher-level embodied interactions including dialogue, reasoning, and decision-making.
Such capabilities are fundamental for service robots, augmented reality assistants, and intelligent embodied systems deployed in real-world environments \cite{song2025hazards,song2025maniplvm}.

Recent advances in large vision language models (LVLMs) have extended the success of 2D perception into the 3D embodied domain, giving rise to unified frameworks that couple perception, language, and action.
As illustrated in Fig.~\ref{fig:main}, these advances lay the foundation for tackling four embodied tasks such as embodied dialogue~\cite{LL3da2024,huang2024chatscene,halacheva2025gaussianvlm}, embodied reasoning~\cite{halacheva2025gaussianvlm,LL3da2024,deepseekr12025}, embodied planning~\cite{octonav2025,LL3da2024,huang2024embodied}, and embodied navigation~\cite{anderson2018vision,krantz2020beyond,zhang2024uninavid,vlnr1}, which are the central focus of this work.

Despite such progress, significant challenges remain.
First, existing approaches often suffer from incoherent and unstable reasoning traces that fail to align with navigation instructions, leading to brittle generalization and semantically inconsistent output.  
Second, embodied navigation requires balancing long-horizon semantic reasoning with low-latency reactive control for real-time execution, a dual requirement that remains largely unaddressed in current methods.

Motivated by these limitations, there is a critical need for embodied foundation models that can jointly address semantic reasoning and embodied action, rather than treating them as separate problems. 
To this end, our work aims to develop a unified framework that balances long-horizon reasoning with real-time responsive control, thereby enabling robust generalization and reliable execution in diverse 3D environments.

To overcome these limitations, we introduce \textbf{\modelname{}}, an embodied foundation model that integrates reasoning, planning, dialogue, and navigation into a unified framework. 
Specifically, we construct a large-scale dataset \textbf{\datasetname{}}, synthesizing high-quality Chains-of-Thought (CoT) for embodied tasks by prompting a strong VLM with egocentric observations, instructions, and action options, followed by rule-based filtering for consistency.
This dataset is used for a cold-start stage that equips \modelname{} with structured reasoning and alignment of instructions.  
Building upon this initialization, we design a reinforcement learning stage with three complementary rewards: (i) a format reward ensuring structural adherence, (ii) an understanding reward capturing semantic correctness and visual grounding, and (iii) a navigation reward optimizing path fidelity and endpoint accuracy.  
To further address the tension between semantic fidelity and real-time control, we propose a Fast-in-Slow dual system reasoning scheme inspired by cognitive science~\cite{Kahneman2011fj}: a slow module aggregates long-term semantics from visual histories, while a fast module executes short-horizon actions with low latency, coordinated asynchronously for robust yet efficient navigation.

We evaluate \modelname{} extensively on the embodied benchmarks.
\modelname{} consistently outperforms prior state-of-the-art methods, achieving improvements in navigation success rates and reasoning accuracy.
Beyond simulation, we deploy \modelname{} on a WHEELTEC R550 mobile robot equipped with a Jetson Orin Nano, LiDAR, and RGB-D camera, demonstrating robust real-world performance under limited onboard computation by designing cloud-based inference.
These results validate that \modelname{} achieves a strong balance of reasoning capability, semantic grounding, and embodied real-time control.

\begin{table}[t]
    \small
    \centering
    \caption{\textbf{Statistics of the public 3D-VL datasets that we draw on when synthesising the \datasetname{} dataset.}
    $N_S$ denotes the scene number.
    \emph{Modality} is the modality within instructions, where $[V,L,P]$ denote $[vision, language, point]$.
    $N_T$ denotes the task number.
    $DE, CE$ denote the discrete and continuous environments.
    }
    \resizebox{\linewidth}{!}{
        \begin{tabular}{l|ccc|ccc|r|c}
        \toprule
        \multirow{2}{*}{Dataset} &  \multicolumn{3}{c|}{Scenes} &  \multicolumn{3}{c|}{Instruction Capability}  &\multirow{2}{*}{$N_T$} & \multirow{2}{*}{Env} \\
        &  MP3D & HM3D & $N_S$ & ObjNav & VLN & Modality & &  \\
        \midrule
        R2R~\cite{anderson2018vision} & \cmark & \xmark & 90 & \xmark & \cmark & $L$ & 22K & DE \\
        R2R-CE~\cite{krantz2020beyond} & \cmark & \xmark & 90 & \xmark & \cmark & $L$ & 4.5K & CE \\
        RxR-CE~\cite{rxr2020} & \cmark & \xmark & 90 & \xmark & \cmark & $L$ & - & CE \\
        SOON~\cite{zhu2021soon} & \cmark & \xmark & 90 & \cmark & \xmark & $L$ & 30K & DE \\
        OVON~\cite{yokoyama2024hm3d} & \xmark & \cmark & 181 & \cmark & \xmark & $L$ & 53K & CE \\ \midrule
        \textbf{\datasetname{} (Ours)}     & \cmark & \cmark & 342 & \cmark & \cmark & $V,L,P$ & 110K & CE \\
        \bottomrule
        \end{tabular}
    }
    \label{tab:stats_dataset}
\end{table}

Our contributions are summarized as follows:
\begin{itemize}
    \item We introduce \textbf{\modelname{}}, an embodied foundation model equipped with a GRPO-based reinforcement learning framework to enhance reasoning and navigation in 3D environments. Specifically, we design three complementary reward functions, \emph{format reward}, \emph{understanding reward}, and \emph{navigation reward}, to improve structural adherence, semantic grounding, and path fidelity. In addition, we construct the large-scale \textbf{\datasetname{}} dataset through a CoT data engine, which provides high-quality step-by-step reasoning trajectories to bootstrap the model via cold-start initialization.
    \item We propose a novel \textbf{Fast-in-Slow reasoning} paradigm that decouples long-horizon semantic reasoning from short-horizon reactive control. And this dual-system design ensures semantic coherence for planning while maintaining low-latency responses in dynamic environments.
    \item We conduct comprehensive evaluations on both embodied AI benchmarks and real-world robot deployment. \modelname{} achieves an average improvement of 8\% compared to strong baselines across dialogue, reasoning, planning, and navigation tasks, and further demonstrates robust performance when deployed on a physical robot with limited on-board resources.

\end{itemize}

\begin{figure*}[t]
    \centering
    \includegraphics[width=\linewidth]{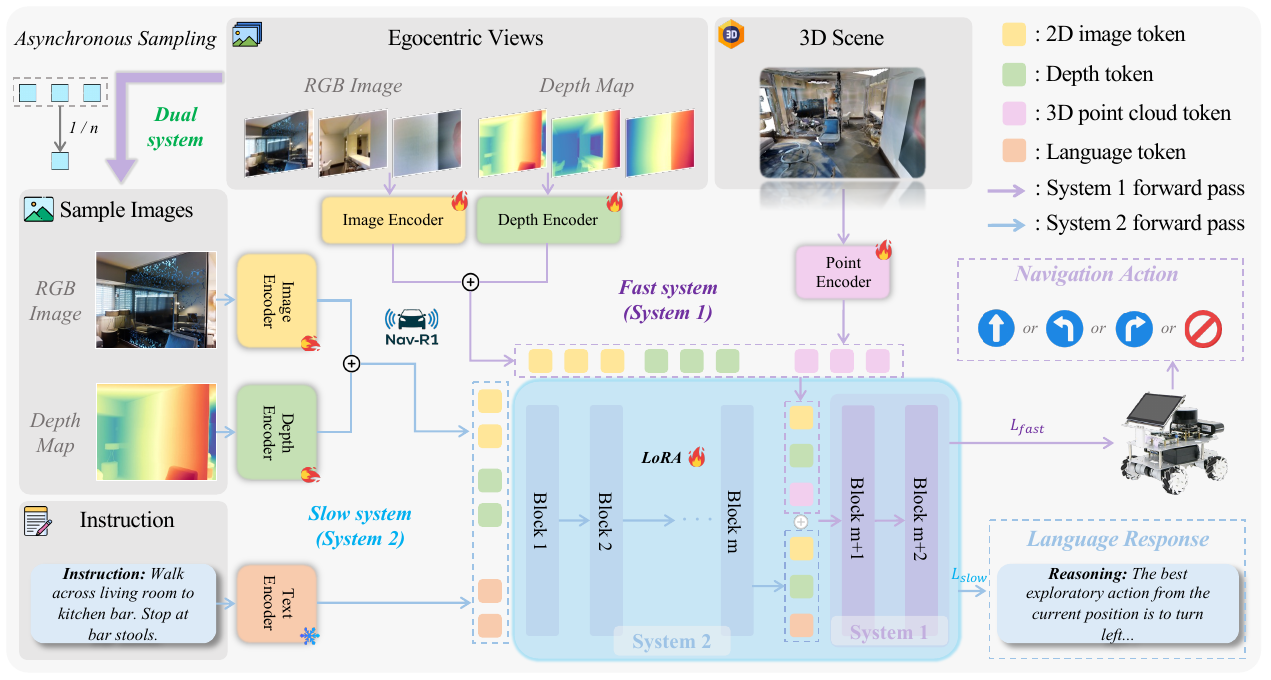}
    \caption{\textbf{Architecture of \modelname{}.} \modelname{} designs a Fast-in-Slow reasoning paradigm that processes egocentric RGB-D views, scene point cloud, and language instructions. The slow system performs long-horizon semantic reasoning, while the fast system executes real-time navigation, enabling coherent reasoning and low-latency control in embodied environments.
    }
    \label{fig:pipeline}
\end{figure*}

\section{Related Work}

\paragraph{Embodied understanding} Embodied understanding seeks to equip agents with the ability to perceive, reason, and act in 3D environments by tightly integrating multimodal sensory data with linguistic instructions.
Early work, limited to 2D visual abstractions, lacked the spatial expressiveness required for complex 3D reasoning.
Recent advances address this gap by using large language models (LLMs) as the connective tissue between perception, grounding, and action planning \cite{huang20253d,huang20253dr1,huang2025dc}. LEO~\cite{huang2024embodied} exemplifies this trend, introducing an embodied generalist agent trained in two stages - 3D vision-language alignment and subsequent vision-language-action instruction tuning, thus achieving unified competence in captioning, question answering and embodied reasoning without task‑specific sub‑modules. Complementing LEO, GaussianVLM~\cite{halacheva2025gaussianvlm} adopts a scene‑centric paradigm that embeds linguistic features directly into 3D Gaussian splats. Its dual sparsification mechanism distills dense scene representations into task‑aware tokens, obviating external object detectors and enabling zero‑shot generalization across embodied reasoning tasks. Together, these studies underscore that unified architectures and language‑aligned 3D representations are crucial to robust scene comprehension and action‑oriented cognition.

\paragraph{Embodied navigation}
Embodied navigation tasks require agents to translate multimodal instructions into smooth, continuous motion through unstructured 3D scenes, which mainly include tasks such as object goal navigation (ObjectNav) and vision language navigation (VLN).
Traditional pipelines rely on discrete topological graphs, which limit path flexibility. Recent research instead embraces end-to-end paradigms powered by large vision language models. VLN‑R1~\cite{vlnr1} couples reinforcement fine-tuning with a time-decayed reward, using GRPO‑style training and the VLN‑Ego dataset to shrink the instruction–action gap.
Uni‑NaVid~\cite{zhang2024uninavid} further unifies the navigation subtasks - object search, instruction follow, and more - within a single video‑conditioned vision‑language‑action backbone, achieving real‑time inference trajectories.
Incoming generalist agents, OctoNav~\cite{octonav2025} introduces a hybrid training paradigm that integrates the chain-of-thought of Think‑Before‑Action with GRPO and RL online, enabling faithful compliance with free‑form commands and continuous control.
Most recently, MTU3D~\cite{zhu2025mtu} bridges visual grounding and active exploration by treating unexplored regions as frontier queries within a unified vision‑language‑exploration objective; pre‑training on one million trajectories lifts success rates by 14–23\% on HM3D‑OVON, GOAT‑Bench, SG3D, and A‑EQA while supporting language, category, and image goals alike. 
Collectively, these advances showcase the growing capacity of LVLMs to fuse perception, reasoning, and action for adaptive embodied navigation.

\section{Datasets}
We build our framework on public 3D vision language datasets and the newly constructed \datasetname{} dataset. 
Table~\ref{tab:stats_dataset} summarizes the statistics of all datasets considered in this work, including scene coverage, task types, instruction modalities, and environment settings.

\subsection{Public Datasets}
We draw on several widely used embodied AI benchmarks to ensure diversity and comparability. 
R2R~\cite{anderson2018vision} and its continuous extension R2R-CE~\cite{krantz2020beyond} provide natural language navigation instructions in Matterport3D scenes. 
RxR-CE~\cite{rxr2020} extends this setup to a multilingual setting with dense temporal grounding. 
For object-goal navigation, SOON~\cite{zhu2021soon} introduces category-conditioned search in indoor environments, while HM3D-OVON~\cite{yokoyama2024hm3d} further supports open-vocabulary object navigation under a zero-shot setting. 
Together, these benchmarks cover both instruction-following and object-centric navigation tasks across discrete and continuous environments.

\subsection{Synthetic Dataset}
Building upon these resources, we introduce \textbf{\datasetname{}}, a large-scale dataset consisting of 110K step-by-step Chain-of-Thought trajectories. 
Unlike prior datasets that primarily provide instructions and target locations, \datasetname{} explicitly includes structured reasoning aligned with multimodal observations, thereby bridging perception, language, and action. 
This dataset serves as the foundation for the cold-start stage of \modelname{}, enabling it to acquire structured reasoning capabilities before reinforcement learning.

\section{The Proposed Method}

\subsection{Overview}
The proposed \textbf{\modelname{}} framework is designed as an embodied foundation model that unifies multimodal perception, structured reasoning, and embodied control.
As illustrated in Fig.~\ref{fig:pipeline}, \modelname{} adopts a Fast-in-Slow reasoning paradigm, where the slow system performs long-horizon semantic reasoning while the fast system ensures low-latency navigation.
The framework follows a two-stage training pipeline: a CoT data engine is first used to construct the \datasetname{} dataset, which provides high-quality step-by-step reasoning trajectories.
Based on this dataset, a cold-start stage initializes the model’s reasoning ability, followed by a reinforcement learning stage with multi-dimensional rewards to refine semantic grounding and navigation fidelity.

\subsection{CoT Data Engine}

We introduce a CoT data engine designed to construct high-quality Chains of Thought (CoT) for embodied navigation and reasoning tasks.
This engine harnesses the reasoning abilities of vision-language models (VLMs) to generate coherent step-by-step rationales that inform navigation decisions in complex 3D environments.

\begin{figure}[t]
    \centering
    \includegraphics[width=\linewidth]{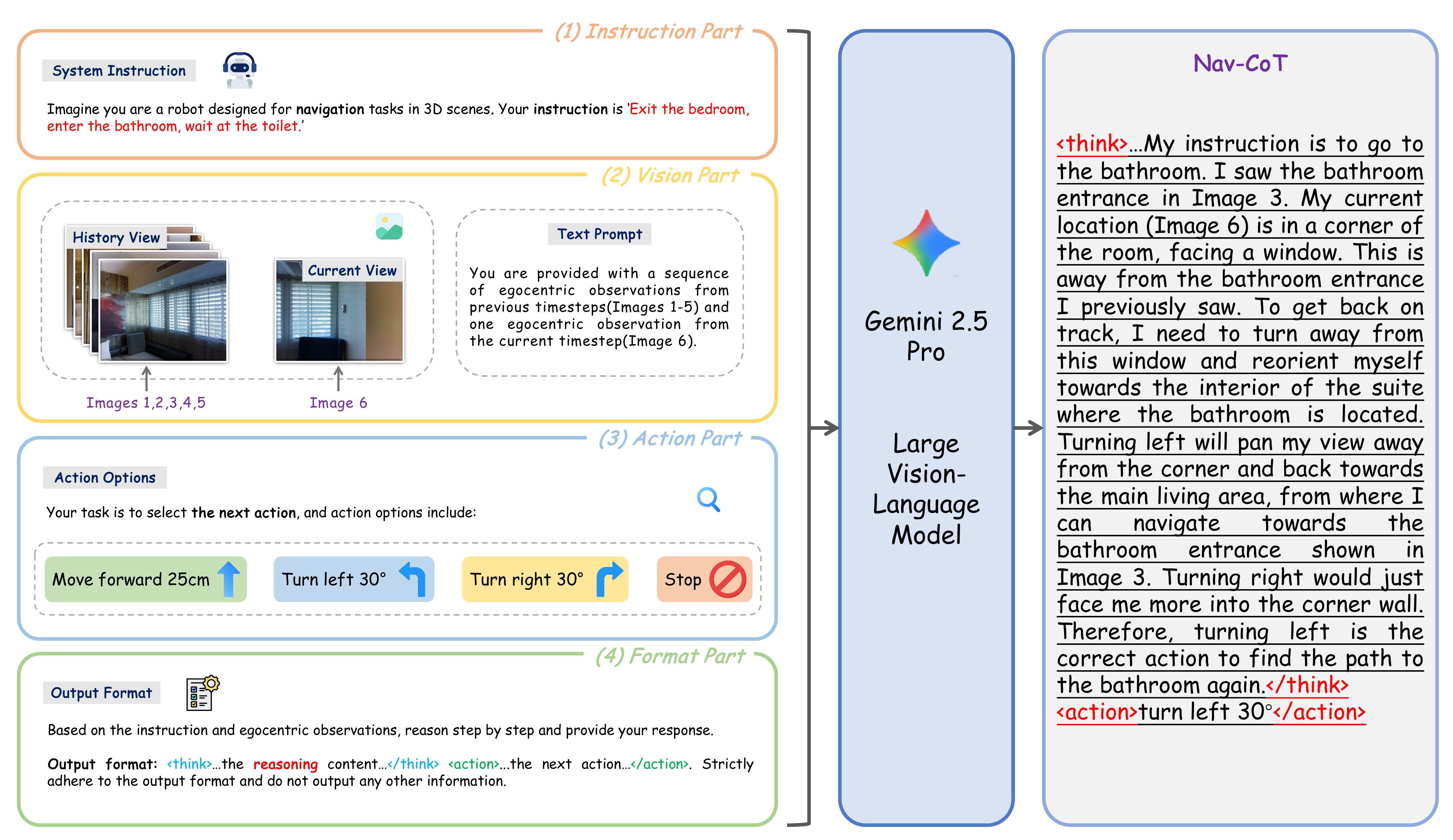}
    \caption{\textbf{CoT Data Engine.} We construct the Nav-CoT dataset by defining navigation instructions, integrating egocentric visual inputs, providing action options and specifying the output format. These components are fed into Gemini 2.5 Pro, which generates step-by-step reasoning and action decisions aligned with navigation goals.}
    \label{fig:cot_data_engine}
\end{figure}

As shown in Fig.~\ref{fig:cot_data_engine}, the process begins with egocentric visual observations extracted from 3D scenes, providing rich contextual views aligned with the agent’s perspective.
In parallel, we incorporate navigation instructions drawn from standard embodied AI benchmarks - VLN tasks rely on R2R~\cite{anderson2018vision}, R2R-CE~\cite{krantz2020beyond}, and RxR-CE~\cite{rxr2020} datasets, while ObjectNav tasks use instructions from SOON~\cite{zhu2021soon} and OVON~\cite{yokoyama2024hm3d}.

To guide reasoning, we design a composite prompt containing four essential components: (1) navigation instruction, (2) egocentric visual input, (3) set of feasible actions at each step, and (4) explicit formatting requirements.
This prompt directs Gemini~2.5 Pro~\cite{geminipro2025} to reason over spatial relations, environment constraints, and instruction semantics, and to produce structured step-by-step CoT sequences.
The outputs follow a standardized format, with reasoning enclosed in \texttt{<think>...</think>} tags and the chosen action in \texttt{<action>...</action>} tags, ensuring transparent alignment between observations, reasoning, and decisions.

Running this pipeline across diverse environments yields approximately 115K CoT examples, each consisting of a scene ID, navigation instruction, visual inputs, structured reasoning, and corresponding action.
These raw outputs are then refined through a two-stage filtering pipeline: (i) rule-based checks to discard incomplete or logically inconsistent responses, and (ii) quality verification by cross-validating actions against feasible navigation paths.
After refinement, the 110K examples form the \datasetname{} dataset, which serves as the cold-start initialization corpus for \modelname{}, providing rich reasoning trajectories that tightly couple perception, instruction following, and navigation decision making.

\subsection{Cold Start Stage}
Although reinforcement learning has shown remarkable effectiveness in reasoning-intensive models such as DeepSeek-R1~\cite{deepseekr12025}, directly applying RL to large 3D vision language models often leads to unstable optimization.
In particular, the policy tends to generate semantically incoherent CoT sequences or produce actions that fail to align with navigation instructions, making it difficult for the model to converge from scratch.

\begin{figure*}[t]
    \centering
    \includegraphics[width=\linewidth]{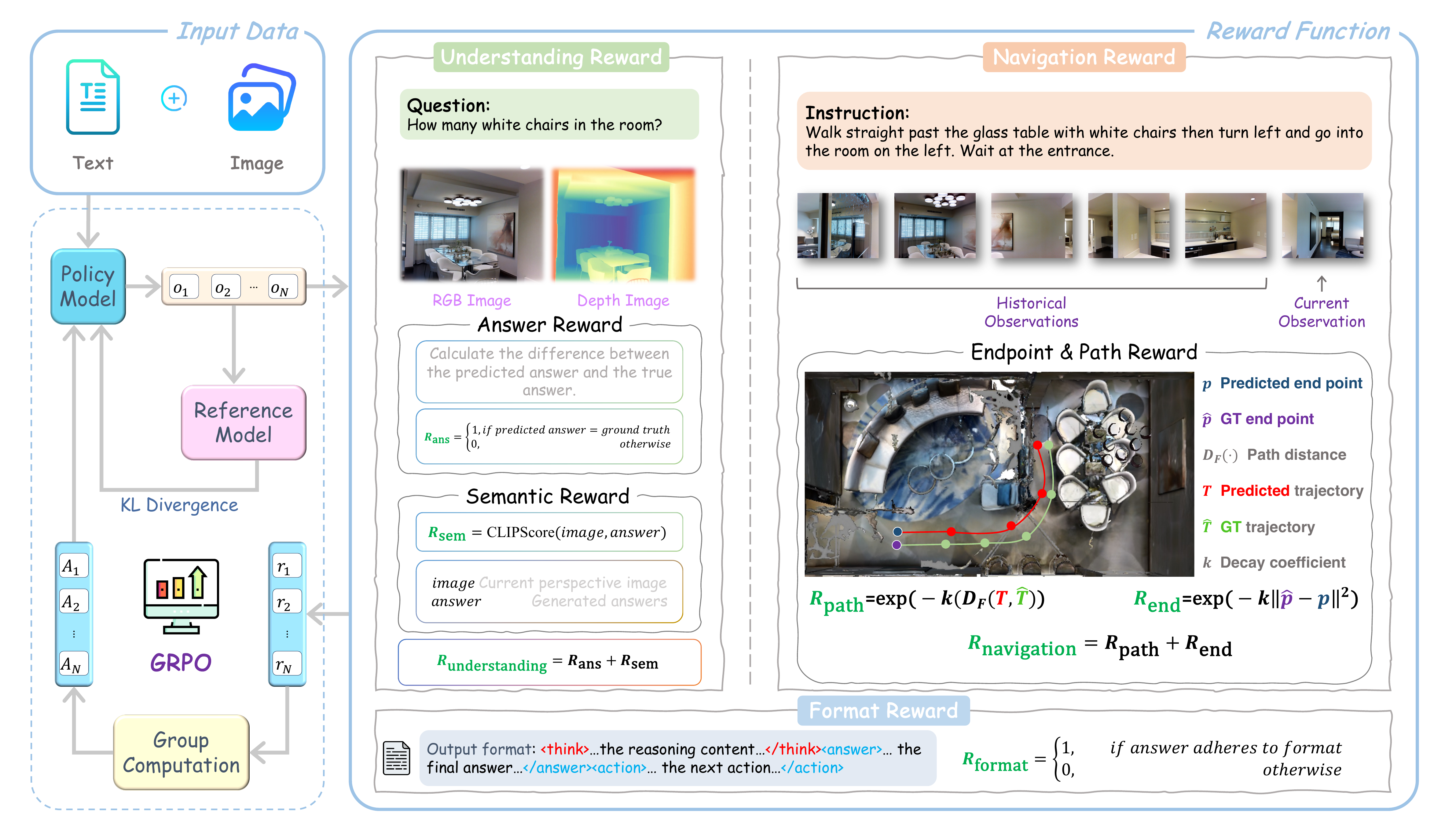}
    \caption{\textbf{The pipeline of RL Policy.} The policy model generates $N$ outputs from text-image input. Then understanding reward (answer correctness and semantic alignment), navigation reward (path fidelity and endpoint accuracy), and format reward (structure adherence) are computed, grouped, and combined with a KL term to a frozen reference model to update the policy.}
    \label{fig:pipeline_rl}
\end{figure*}

To address this issue, we adopt a cold-start stage based on supervised fine-tuning. Specifically, the \datasetname{} dataset generated by our CoT data engine is employed to initialize \modelname{}.
This supervised training step equips the policy with essential capabilities to produce structured reasoning sequences in the format of \texttt{<think>...</think><action>...</action>} and ground them to corresponding navigation actions.

By bootstrapping the model with coherent reasoning and action patterns, the cold-start stage stabilizes subsequent RL optimization and provides a smooth transition to the reinforcement learning stage, where multi-dimensional rewards further refine semantic understanding, path fidelity, and structural adherence.

\subsection{Reinforcement Learning}
Group Relative Policy Optimization (GRPO)~\cite{deepseek-math} has recently demonstrated strong effectiveness in reasoning-intensive tasks, such as DeepSeek R1~\cite{deepseekr12025}. 
Its core principle is to refine the policy through \emph{group-based feedback}: multiple candidate responses are sampled from the current policy, scored by task-specific reward functions, and updated according to their relative advantages.

As depicted in Fig.~\ref{fig:pipeline_rl}, extending this framework to embodied 3D vision-language tasks, \modelname{} introduces three complementary rewards that jointly supervise training: \emph{Format Reward}, which enforces the structural validity of the outputs; \emph{Understanding Reward}, which ensures semantic correctness and visual grounding in 3D scene reasoning; and \emph{Navigation Reward}, which evaluates path fidelity and goal-reaching accuracy in navigation.
Together, these rewards provide multidimensional feedback that balances linguistic structure, semantic understanding, and action execution.

\paragraph{Policy samples} For an input state $(x, q)$, where $x$ encodes multi-modal scene representations and $q$ denotes the instruction or question, \modelname{} generates $N$ candidate responses $\{o_1, o_2, \cdots, o_N\}$ from policy $\pi_\theta$.  
Each candidate corresponds to either an answer prediction for scene understanding or an action prediction for navigation.  
These responses are then evaluated using the following reward functions.  

\paragraph{Format reward} To guarantee structural consistency, \emph{format reward} $R_{\text{Format}}$ verifies whether each output strictly adheres to the reasoning-decision template: \texttt{<think>...</think><answer>...</answer>} or \texttt{<think>...</think><action>...</action>}.
Formally,
\begin{equation}
    R_{\text{Format}} =  \left\{\begin{matrix} 
    1 ,& \text{if output adheres to format}\\ 
    0 ,& \text{otherwise} 
    \end{matrix}\right. .
\end{equation}
This constraint not only ensures machine-parseable outputs but also disentangles reasoning from final predictions.

\paragraph{Understanding reward} The \emph{understanding reward} $R_{\text{understanding}}$ assesses whether the model demonstrates genuine comprehension of the 3D scene.
It consists of two components: \emph{Answer Reward} and \emph{Semantic Reward}.
The answer reward measures exact correctness against ground truth:
\begin{equation}
    R_{\text{ans}} = \left\{\begin{matrix} 
    1 ,& \text{if predicted answer equals ground truth}\\ 
    0 ,& \text{otherwise} 
    \end{matrix}\right. ,
\end{equation}
while the semantic reward measures alignment between the generated answer $\hat{a}$ and the paired RGB-D image $I$:
\begin{equation}
    R_{\text{sem}} = \mathrm{CLIPScore}(I, \hat{a}).
\end{equation}
The overall understanding reward is defined as:
\begin{equation}
    R_{\text{understanding}} = R_{\text{ans}} + R_{\text{sem}}.
\end{equation}
This design prevents both factual errors and semantically irrelevant outputs.

\paragraph{Navigation reward} For navigation tasks, the \emph{navigation Reward} $R_{\text{navigation}}$ evaluates whether the agent follows the instruction and successfully reaches the target.  
It integrates two components: a \emph{path reward}, which measures the fidelity of the trajectory with respect to the reference path $\hat{T}$, and an \emph{endpoint reward}, which enforces the precision of the final location $\hat{p}$.

Given a predicted trajectory $T$ and a ground-truth trajectory $\hat{T}$, the path reward is defined as:
\begin{equation}
    R_{\text{path}} = \exp\big(-k D_F(T, \hat{T})\big),
\end{equation}
where $D_F(\cdot)$ denotes a trajectory distance metric and $k$ is a decay coefficient.

Furthermore, the endpoint reward penalizes deviations between the predicted endpoint $p$ and the ground-truth endpoint $\hat{p}$:
\begin{equation}
    R_{\text{end}} = \exp\big(-k \lVert \hat{p} - p \rVert^2\big).
\end{equation}
The final navigation reward is the combination of both:
\begin{equation}
    R_{\text{navigation}} = R_{\text{path}} + R_{\text{end}}.
\end{equation}
This formulation ensures that both trajectory-level alignment and goal-reaching ability are optimized.

\paragraph{Policy optimization} Inspired by Group Relative Policy Optimization (GRPO)~\cite{deepseek-math}, we sample multiple candidate responses $\{o_1, o_2, \cdots, o_N\}$ from the current policy $\pi_\theta$, and compute their corresponding rewards $\mathbf{r} = \{r_1, r_2, \cdots, r_N\}$ from the above functions.
Each reward is normalized to compute the relative advantage:
\begin{equation}
    \hat{A}_i = \frac{r_i - \mathrm{mean}(\mathbf{r})}{\mathrm{std}(\mathbf{r})},
\end{equation}
where $\hat{A}_i$ denotes the advantage of the $i$-th response.
The policy is then updated by maximizing the clipped GRPO objective with KL regularization:
\begin{equation}
\allowdisplaybreaks
\begin{aligned}
\mathcal{J}_{\mathrm{GRPO}}(\theta) 
= &\mathbb{E}_{c}
\Biggl[ \frac{1}{G}\sum_{i=1}^G \biggl( 
\min\left(\frac{\pi_\theta(o_i|q)}{\pi_{\theta_{\mathrm{old}}}(o_i|q)}\hat{A}_i,\right.\\[5pt]
&\quad\left.\mathrm{clip}\left(
\frac{\pi_\theta(o_i|q)}{\pi_{\theta_{\mathrm{old}}}(o_i|q)},
1-\varepsilon,
1+\varepsilon
\right)\hat{A}_i\right)\\[5pt]
&\quad -\beta \cdot \mathbb{D}_{\mathrm{KL}}(\pi_\theta\|\pi_{\mathrm{ref}})\biggr)\Biggr].
\end{aligned}
\end{equation}

\subsection{Fast-in-Slow Reasoning}
Inspired by dual system theories of human cognition~\cite{Kahneman2011fj}, we propose a \textbf{Fast-in-Slow} paradigm for \modelname{}, which tightly couples deliberate semantic reasoning with rapid action execution.
As shown in Fig.~\ref{fig:pipeline}, this design addresses the tension between accurate long-horizon planning and low-latency control in dynamic embodied environments.

\begin{table}[t]
\centering
\caption{\textbf{Object goal navigation results on HM3D-OVON~\cite{yokoyama2024hm3d}.} $^*$ denotes zero-shot methods.}
\resizebox{\linewidth}{!}{
\begin{tabular}{l|cc|cc|cc}
\toprule
\multirow{2}{*}{Method} & \multicolumn{2}{c|}{Val-Seen} & \multicolumn{2}{c|}{Val-Seen-Synonyms} & \multicolumn{2}{c}{Val-Unseen} \\
\cmidrule(lr){2-3} \cmidrule(lr){4-5} \cmidrule(lr){6-7}
& \textbf{SR}$\uparrow$ & \textbf{SPL}$\uparrow$ & \textbf{SR}$\uparrow$ & \textbf{SPL}$\uparrow$ & \textbf{SR}$\uparrow$ & \textbf{SPL}$\uparrow$ \\
\midrule
BC~\cite{alvinn1988}             & 11.1 & 4.5  & 9.9  & 3.8  & 5.4  & 1.9 \\
DAgger~\cite{dagger2011}         & 11.1 & 4.5  & 9.9  & 3.8  & 5.4  & 1.9 \\
RL~\cite{ppo2017}                & 18.1 & 9.4  & 15.0 & 7.4  & 10.2 & 4.7 \\
BCRL~\cite{rcsvln2019}           & 39.2 & 18.7 & 27.8 & 11.7 & 18.6 & 7.5 \\
DAgRL~\cite{touchdown2019}       & 41.3 & 21.2 & 29.4 & 14.4 & 18.3 & 7.9 \\
VLFM$^*$~\cite{yokoyama2024vlfm} & 35.2 & 18.6 & 32.4 & 17.3 & 35.2 & 19.6 \\
DAgRL+OD~\cite{yokoyama2024hm3d} & 38.5 & 21.1 & 39.0 & 21.4 & 37.1 & 19.8 \\
Uni-NaVid$^*$~\cite{zhang2024uninavid} & 41.3 & 21.1 & 43.9 & 21.8 & 39.5 & 19.8 \\
MTU3D$^*$~\cite{zhu2025mtu}      & 55.0 & 23.6 & 45.0 & 14.7 & 40.8 & 12.1 \\ \midrule
\textbf{\modelname{} (Ours)} & \textbf{58.4} & \textbf{26.3} & \textbf{48.1} & \textbf{23.1} & \textbf{42.2} & \textbf{20.1} \\
\bottomrule
\end{tabular}
}
\label{tab:objnav}
\end{table} 

\paragraph{Slow reasoning} The slow system (System~2) operates at a lower frequency and processes multimodal observations, including egocentric RGB-D frames and language instructions.
It aggregates the historical context into compact memory states and outputs latent features $h_t$ that encode scene semantics, temporal dependencies, and global navigation goals.
These structured features provide high-level guidance to ensure coherent decision-making. In practice, the slow system aggregates visual history into compact memory states, enabling \modelname{} to maintain semantic consistency at the scene level while avoiding excessive computation.

\begin{table*}[t]
    \centering
    \caption{\textbf{Comparison with state-of-the-art methods on the Val-Unseen split of R2R-CE~\cite{anderson2018vision} and RxR-CE~\cite{rxr2020}.} $^{*}$ indicates methods using the waypoint predictor from~\cite{hong2022bridging}. \modelname{} outperforms all methods that do not rely on simulator pre-trained waypoint predictors, even when those methods leverage additional inputs such as depth, panoramic views, and odometry.}
    \resizebox{\textwidth}{!}{
    \begin{tabular}{lcccclcccclcccc}
    \toprule
    \multirow{2}[2]{*}{Method} & \multicolumn{4}{c}{Observation} & & \multicolumn{4}{c}{R2R-CE Val-Unseen} & & \multicolumn{4}{c}{RxR-CE Val-Unseen} \\
    \cmidrule(lr){2-5} \cmidrule(lr){7-10} \cmidrule(lr){12-15}
    & S.RGB & Pano. & Depth & Odom. & & NE$\downarrow$ & OS$\uparrow$ & SR$\uparrow$ & SPL$\uparrow$ & & NE$\downarrow$ & SR$\uparrow$ & SPL$\uparrow$ & nDTW$\uparrow$ \\
    \midrule
    CMA$^{*}$~\cite{hong2022bridging}           & \xmark & \cmark & \cmark & \cmark & & 6.20 & 52.0 & 41.0 & 36.0 & & 8.76 & 26.5 & 22.1 & 47.0 \\
    Sim2Sim$^{*}$~\cite{krantz2022sim}          & \xmark & \cmark & \cmark & \cmark & & 6.07 & 52.0 & 43.0 & 36.0 & & - & - & - & - \\
    GridMM$^{*}$~\cite{wang2023gridmm}          & \xmark & \cmark & \cmark & \cmark & & 5.11 & 61.0 & 49.0 & 41.0 & & - & - & - & - \\
    Ego$^{2}$-Map$^{*}$~\cite{hong2023learning} & \xmark & \cmark & \cmark & \cmark & & 5.54 & 56.0 & 47.0 & 41.0 & & - & - & - & - \\
    DreamWalker$^{*}$~\cite{wang2023dreamwalker}& \xmark & \cmark & \cmark & \cmark & & 5.53 & 59.0 & 49.0 & 44.0 & & - & - & - & - \\
    Reborn$^{*}$~\cite{an20221st}               & \xmark & \cmark & \cmark & \cmark & & 5.40 & 57.0 & 50.0 & 46.0 & & 5.98 & 48.6 & 42.0 & 63.3 \\
    ETPNav$^{*}$~\cite{an2024etpnav}            & \xmark & \cmark & \cmark & \cmark & & 4.71 & 65.0 & 57.0 & 49.0 & & 5.64 & 54.7 & 44.8 & 61.9 \\
    HNR$^{*}$~\cite{wang2024lookahead}          & \xmark & \cmark & \cmark & \cmark & & 4.42 & 67.0 & 61.0 & 51.0 & & 5.50 & 56.3 & 46.7 & 63.5 \\
    \midrule
    AG-CMTP~\cite{chen2021topological}          & \xmark & \cmark & \cmark & \cmark & & 7.90 & 39.0 & 23.0 & 19.0 & & - & - & - & - \\
    R2R-CMTP~\cite{chen2021topological}         & \xmark & \cmark & \cmark & \cmark & & 7.90 & 38.0 & 26.0 & 22.0 & & - & - & - & - \\
    InstructNav~\cite{long2024instructnavzeroshotgenericinstruction} & \xmark & \cmark & \cmark & \cmark & & 6.89 & - & 31.0 & 24.0 & & - & - & - & - \\
    LAW~\cite{raychaudhuri2021language}         & \cmark & \xmark & \cmark & \cmark & & 6.83 & 44.0 & 35.0 & 31.0 & & 10.90 & 8.0 & 8.0 & 38.0 \\
    CM2~\cite{georgakis2022cross}               & \cmark & \xmark & \cmark & \cmark & & 7.02 & 41.0 & 34.0 & 27.0 & & - & - & - & - \\
    WS-MGMap~\cite{chen2022weakly}              & \cmark & \xmark & \cmark & \cmark & & 6.28 & 47.0 & 38.0 & 34.0 & & - & - & - & - \\
    AO-Planner~\cite{chen2024affordances}       & \xmark & \cmark & \cmark & \xmark & & 5.55 & 59.0 & 47.0 & 33.0 & & 7.06 & 43.3 & 30.5 & 50.1 \\
    Seq2Seq~\cite{krantz2020beyond}             & \cmark & \xmark & \cmark & \xmark & & 7.77 & 37.0 & 25.0 & 22.0 & & 12.10 & 13.9 & 11.9 & 30.8 \\
    CMA~\cite{krantz2020beyond}                 & \cmark & \xmark & \cmark & \xmark & & 7.37 & 40.0 & 32.0 & 30.0 & & - & - & - & - \\
    NaVid~\cite{zhang2024navid}                 & \cmark & \xmark & \xmark & \xmark & & 5.47 & 49.0 & 37.0 & 35.0 & & - & - & - & - \\
    Uni-NaVid~\cite{zhang2024uninavid}          & \cmark & \xmark & \xmark & \xmark & & 5.58 & 53.5 & 47.0 & 42.7 & & 6.24 & 48.7 & 40.9 & - \\
    NaVILA~\cite{cheng2024navila}               & \cmark & \xmark & \xmark & \xmark & & 5.22 & 62.5 & 54.0 & 49.0 & & 6.77 & 49.3 & 44.0 & 58.8 \\
    VLN-R1~\cite{vlnr1}                         & \cmark & \xmark & \xmark & \xmark & & 7.00 & 41.2 & 30.2 & 21.8 & & 9.10 & 22.7 & 17.6 & -    \\
    OctoNav~\cite{octonav2025}                  & \cmark & \xmark & \xmark & \xmark & & -    & 42.9 & 37.1 & 33.6 & & -    & -    & -    & -    \\
    StreamVLN~\cite{streamvln}                  & \cmark & \xmark & \xmark & \xmark & & 4.98 & 64.2 & 56.9 & 51.9 & & 6.22 & 52.9 & 46.0 & 61.9 \\
    CorrectNav~\cite{correctnav2025}            & \cmark & \xmark & \xmark & \xmark & & 4.24 & 67.5 & 65.1 & 62.3 & & 4.09 & 69.3 & 63.3 & 75.2 \\
    \midrule
    \textbf{\modelname{} (Ours)}                & \cmark & \xmark & \cmark & \xmark & & \textbf{3.86} & \textbf{74.1} & \textbf{72.5} & \textbf{68.8} & & \textbf{3.98} & \textbf{71.3} & \textbf{66.3} & \textbf{79.4} \\
\bottomrule
\end{tabular}
}
\label{tab:embodied_navigation}
\end{table*}

\paragraph{Fast reasoning} The fast system (System~1) runs at a higher frequency and ensures real-time responsiveness in dynamic environments.
Instead of independently reasoning, it reuses the final transformer blocks of \modelname{} to inherit pretrained knowledge from System~2 while keeping its computation lightweight.
At each step, the fast system fuses high-frequency egocentric multimodal inputs, including RGB frames, depth maps, and point cloud tokens $(o_{t+1}, \dots, o_{t+H})$, with the latent feature $h_t$ from the slow system to predict a short-horizon sequence of actions:
\begin{equation}
    \{a_{t+1}, \dots, a_{t+H}\} = \pi_{\text{fast}}(o_{t+1:t+H}, h_t),
\end{equation}
where $\pi_{\text{fast}}$ denotes the high-frequency policy model that integrates visual, depth, and 3D geometric cues for real-time control.

\paragraph{Asynchronous coordination} To balance efficiency and accuracy, we design an asynchronous update mechanism with a frequency ratio of $1\!:\!n$ between the slow and fast systems.
An update from the slow system provides latent guidance for $n$ consecutive steps of fast execution.
This decoupling strategy ensures that global semantics remain stable, while local control is executed with low latency.
Empirically, we find that $n \approx 3$ achieves the best balance between semantic fidelity and responsiveness in embodied navigation tasks, yielding both robust long-horizon reasoning and efficient real-time control.

\section{Experiments}

\subsection{Benchmarks and Metrics}
\paragraph{Benchmarks} To thoroughly evaluate \modelname{}, we adopt a diverse set of embodied AI benchmarks that span navigation, dialogue, reasoning, and planning tasks.
For embodied navigation, we benchmark on R2R-CE~\cite{krantz2020beyond} and RxR-CE~\cite{rxr2020}, where agents navigate to goal locations in unseen environments, as well as HM3D for standard object goal navigation. To further test open-vocabulary generalization, we adopt HM3D-OVON~\cite{yokoyama2024hm3d}, which introduces novel categories under a zero-shot setting.
Embodied dialogue and planning are evaluated on 3D-LLM~\cite{hong2023dllm}, which requires generating natural responses and coherent multi-step action plans.
Embodied reasoning is assessed on SQA3D~\cite{ma2022sqa3d}, which involves spatial question answering over complex 3D scenes.

\paragraph{Metrics} For embodied navigation, we follow standard metrics~\cite{anderson2018vision} including the navigation error (NE), success rate (SR), oracle success rate (OS), success weighted by path length (SPL)~\cite{anderson2018spl}, and normalized dynamic time warping (nDTW)~\cite{ndtw}.
For embodied dialogue, planning, and reasoning tasks, we report widely used language generation metrics such as CIDEr (C)~\cite{cider2015}, BLEU-4 (B-4)~\cite{bleu2002}, METEOR (M)~\cite{meteor2005}, and ROUGE-L (R)~\cite{rouge2004}.

\subsection{Implementation Details}

\begin{table*}[t]
\centering
\caption{\textbf{Embodied dialogue and planning} results on 3D-LLM~\cite{hong2023dllm}. \textbf{Embodied reasoning} results on SQA3D~\cite{ma2022sqa3d}.}
\resizebox{\linewidth}{!}{
\begin{tabular}{lcccccccccccc}
\toprule
\multirow{2}[2]{*}{Method} & \multicolumn{4}{c}{\textbf{Embodied dialogue}} & \multicolumn{4}{c}{\textbf{Embodied planning}} & \multicolumn{4}{c}{\textbf{Embodied reasoning}} \\
\cmidrule(lr){2-5} \cmidrule(lr){6-9} \cmidrule(lr){10-13}
 & C$\uparrow$ & B-4$\uparrow$ & M$\uparrow$ & R$\uparrow$ & C$\uparrow$ & B-4$\uparrow$ & M$\uparrow$ & R$\uparrow$ & C$\uparrow$ & B-4$\uparrow$ & M$\uparrow$ & R$\uparrow$ \\
\midrule
LL3DA~\cite{LL3da2024}             & 190.01 & 23.95 & 23.50 & 40.61 & 128.80 & 12.95 & 17.05 & 39.25 & - & - & - & - \\
Spatial 3D-LLM~\cite{spatial3dllm} & - & - & - & - & 195.92 & 14.65 & 18.95 & 36.93 & -  & - & -  & - \\
LSceneLLM~\cite{zhi2024lscenellm}  & 104.98 & - & 21.26 & 36.00 & 214.63 & - & 21.05 & 47.05 & - & - & - & - \\
LEO~\cite{huang2024an}             & - & - & - & - & - & - & - & - & 124.70 & 9.40 & 25.50 & 48.40 \\
3D-R1~\cite{huang20253dr1}         & 280.34 & \textbf{39.45} & 66.89 & \textbf{55.34} & 230.50 & 25.45 & \textbf{48.34} & 55.67 & 138.67 & \textbf{23.56} & 35.45 & \textbf{60.02} \\
\midrule
\textbf{\modelname{} (Ours)}       & \textbf{281.20} & 39.34 & \textbf{67.53} & 55.12 & \textbf{230.52} & \textbf{25.98} & 47.11 & \textbf{56.23} & \textbf{139.98} & 23.20 & \textbf{36.15} & 59.50 \\
\bottomrule
\end{tabular}
}
\label{exp:comparison_on_dialoguereasoningplanning}
\end{table*}

\paragraph{Data synthesis} We first construct the \datasetname{} dataset using our CoT data engine.
The instructions are sampled from R2R~\cite{anderson2018vision}, R2R-CE~\cite{krantz2020beyond}, RxR-CE~\cite{rxr2020}, SOON~\cite{zhu2021soon}, and HM3D-OVON~\cite{yokoyama2024hm3d}.
For each scene, we provide RGB-D egocentric input, candidate action sets, and explicit output formatting to Gemini~2.5 Pro~\cite{geminipro2025}, which generates reasoning traces in \texttt{<think>} tags and the corresponding actions in \texttt{<action>} tags or answers in \texttt{<answer>} tags.
A two-stage filtering process is applied: (i) rule-based checks discard incomplete or logically inconsistent responses, and (ii) trajectory verification ensures action feasibility against ground-truth paths. After filtering, 110K high-quality trajectories remain and are used for cold-start training.

\paragraph{Cold-start initialization} We initialize \modelname{} from the pre-trained 3D-R1 model~\cite{huang20253dr1}, which already provides strong 3D reasoning and vision-language alignment.
On top of this initialization, we perform supervised fine-tuning (SFT) on \datasetname{} for 2 epochs with a batch size of 8. The AdamW optimizer is used with weight decay $0.01$, and a cosine annealing learning rate schedule decays from $10^{-4}$ to $10^{-5}$.
This stage equips \modelname{} with the ability to generate coherent reasoning-action sequences of the form \texttt{<think>...</think><action>...</action>}, ensuring structural adherence and semantic grounding before reinforcement learning.

\paragraph{Reinforcement learning} After cold-start initialization, we fine-tune the model with Group Relative Policy Optimization (GRPO)~\cite{deepseek-math}.
For each input, multiple responses are sampled and scored with three complementary rewards: (i) a format reward enforcing structural validity, (ii) an understanding reward that combines exact match correctness with CLIP-based semantic alignment, and (iii) a navigation reward measuring trajectory fidelity and endpoint accuracy.
RL training runs for 2 epochs with batch size 12, fixed learning rate $10^{-5}$, and a KL penalty $\beta=0.02$ against the frozen SFT policy.

\paragraph{Parameter efficient tuning} To reduce training cost, we adopt parameter-efficient fine-tuning by injecting LoRA adapters~\cite{hu2022lora} into the last 8 transformer blocks of the backbone.
Each adapter is configured with rank $r=6$ and scaling factor $\alpha=8$, introducing $\sim$12M trainable parameters.
In total, about 142M parameters are updated, reducing trainable parameters by $\sim$98\% compared to full fine-tuning.
All experiments are conducted on 4$\times$NVIDIA H20 GPUs.

\begin{table}[t]
\centering
\caption{\textbf{Quantitative results of real-world experiments across three distinct indoor environments.} We report navigation error (NE) and success rate (SR) for \modelname{} and four baselines. \modelname{} consistently outperforms all competing methods.
}
\resizebox{\linewidth}{!}{
\begin{tabular}{lcccccc}
\toprule
\multirow{2}[2]{*}{Method} & \multicolumn{2}{c}{\texttt{Meeting Room}} & \multicolumn{2}{c}{\texttt{Lounge}} & \multicolumn{2}{c}{\texttt{Corridor}} \\
\cmidrule(lr){2-3} \cmidrule(lr){4-5} \cmidrule(lr){6-7}
 & NE$\downarrow$ & SR$\uparrow$ & NE$\downarrow$ & SR$\uparrow$ & NE$\downarrow$ & SR$\uparrow$ \\
\midrule
NaVILA~\cite{cheng2024navila}         & 2.06 & 0.45 & 2.21 & 0.45 & 1.97 & 0.50  \\
NaVid~\cite{zhang2024navid}           & 1.88 & 0.55 & 2.22 & 0.50 & 1.94 & 0.55  \\
Uni-NaVid~\cite{zhang2024uninavid}    & 1.76 & 0.64 & 2.13 & 0.64 & 1.87 & 0.65  \\
MTU3D~\cite{zhu2025mtu}               & 1.64 & 0.73 & 1.98 & 0.81 & 1.62 & 0.70  \\
\midrule
\textbf{\modelname{} (Ours)}          & \textbf{1.23} & \textbf{1.03} & \textbf{0.98} & \textbf{1.12} & \textbf{1.24} & \textbf{1.02} \\
\bottomrule
\end{tabular}
}
\label{tab:real_eval}
\end{table}

\subsection{Main Results}

\paragraph{Embodied dialogue} Table~\ref{exp:comparison_on_dialoguereasoningplanning} shows that \modelname{} maintains strong dialogue capability, achieving results close to 3D-R1 while outperforming previous baselines.
This indicates that incorporating navigation-oriented reasoning does not weaken interaction quality.

\paragraph{Embodied reasoning} For embodied reasoning, as shown in Table~\ref{exp:comparison_on_dialoguereasoningplanning}, \modelname{} performs comparably to 3D-R1.
This is consistent with our design choice, as we do not train additional understanding modules but instead preserve the same scene reasoning ability while prioritizing navigation improvements.

\paragraph{Embodied planning} For embodied planning, the results in Table~\ref{exp:comparison_on_dialoguereasoningplanning} indicate that \modelname{} performs on par with prior methods, generating coherent multi-step action sequences.
This confirms that our Fast-in-Slow design and GRPO training preserve planning skills while primarily optimizing navigation.

\paragraph{Embodied navigation} As shown in Table~\ref{tab:embodied_navigation} and Table~\ref{tab:objnav}, \modelname{} consistently outperforms prior methods on both instruction-following and object-goal navigation benchmarks.
It achieves higher success rates and trajectory efficiency while reducing navigation errors, demonstrating superior generalization across unseen environments.

\subsection{Real World Evaluation}
\begin{figure}[t]
    \centering
    \includegraphics[width=\linewidth]{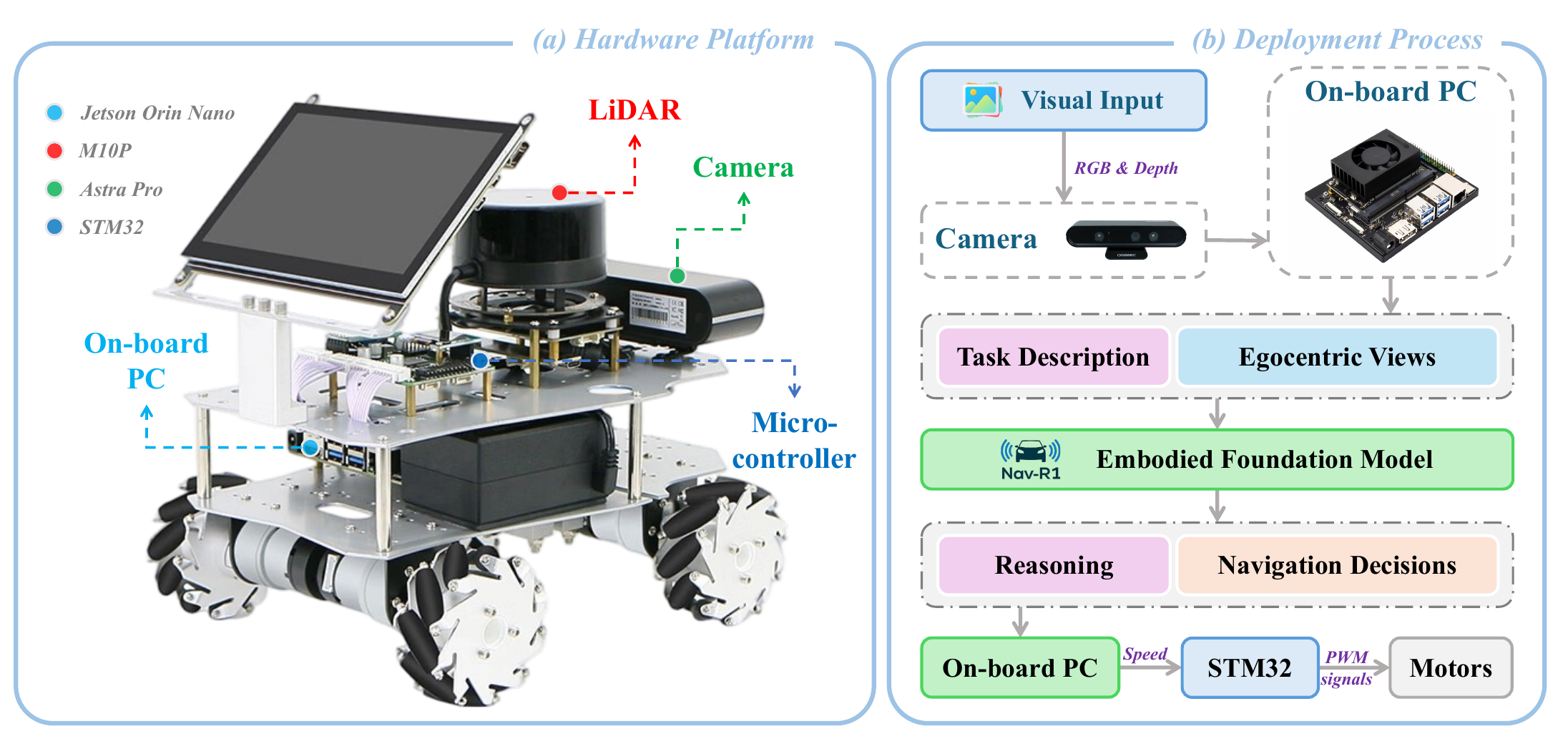}
    \caption{\textbf{Real-world robot setup and deployment pipeline.}
    (a) Hardware platform: the WHEELTEC R550 robot equipped with Jetson Orin Nano (on-board PC), M10P LiDAR for mapping, Astra Pro RGB-D camera for perception, and STM32 microcontroller for motor control.
    (b) Deployment process: egocentric visual inputs are transmitted to the embodied foundation model \modelname{}, which performs reasoning and navigation. The decisions are then sent back to the on-board PC and converted into low-level motor commands by the STM32 controller.
    }
    \label{fig:robot_setup}
\end{figure}

\begin{figure*}[t]
    \centering
    \includegraphics[width=\linewidth]{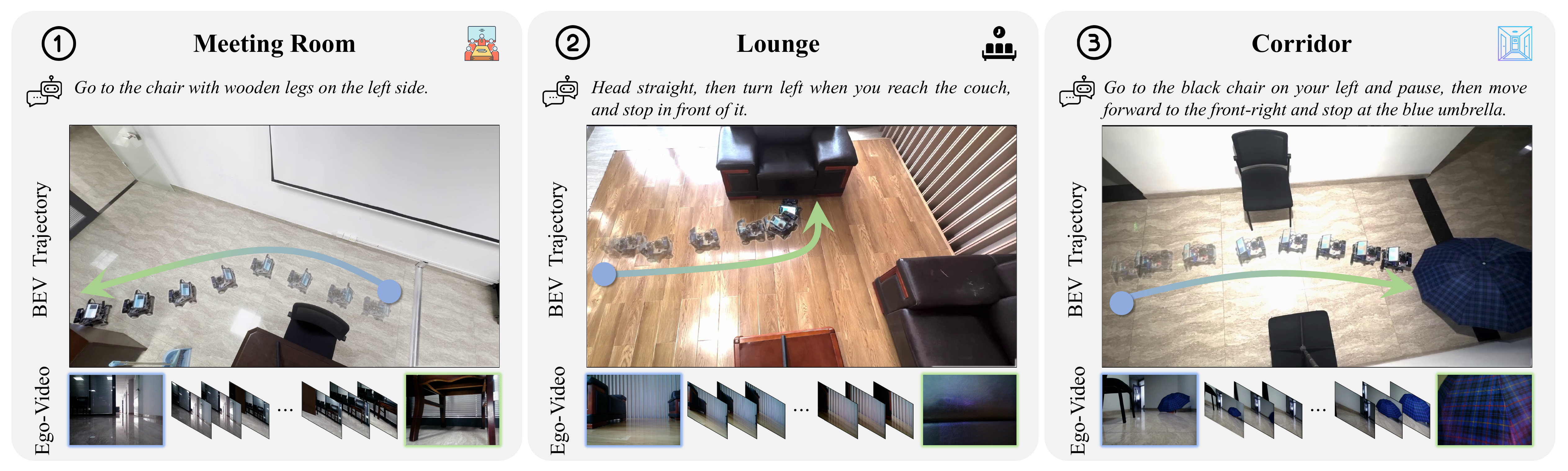}
    \caption{\textbf{Qualitative results from the real-world deployment of \modelname{}.} We evaluate the agent in three indoor scenarios: meeting room, lounge, and corridor. Each scene illustrates the BEV trajectory and ego-centric video frames, showing the model’s ability to generalize to diverse layouts and object configurations in real-world environments.
    }
    \label{fig:realworld_vis}
    \vspace{-0.6cm}
\end{figure*}
\paragraph{Robot type settings} As shown in Fig.~\ref{fig:robot_setup}, we use the WHEELTEC R550 as the mobile platform for real-world evaluation.
The robot is equipped with a Jetson Orin Nano as the on-board computing unit, an M10P LiDAR for environmental mapping, and an Astra Pro camera for RGB-D perception.
Considering the limited edge computing resources of the platform, the embodied foundation model \modelname{} is deployed on a cloud server rather than running locally.
The system operates in a closed loop manner: the robot transmits egocentric RGB inputs to the cloud, where \modelname{} performs reasoning and generates navigation decisions.
These commands are then transmitted back to the on-board system, where an STM32 microcontroller converts them into PWM signals that directly control the robot’s motors.

\paragraph{Scene setup and task types} To assess generalizability under various real-world conditions, we evaluate \modelname{} in three distinct indoor scenes: \texttt{a meeting room}, \texttt{a lounge}, and \texttt{a corridor}, each characterized by unique spatial layouts and object distributions.
Across these settings, the robot is instructed to perform navigation-oriented tasks that vary in difficulty, ranging from short-horizon paths with clear line of sight to long-horizon trajectories involving clutter, obstacles, and occlusions.
This setup enables a comprehensive assessment of the robustness of \modelname{} in handling heterogeneous layouts and task complexities.

\paragraph{Quantitative real-world evaluation} To quantify performance, we compare \modelname{} against previous navigation models including NaVILA~\cite{cheng2024navila}, NaVid~\cite{zhang2024navid}, Uni-NaVid~\cite{zhang2024uninavid}, and MTU3D~\cite{zhu2025mtu}.
Each model is tested in the meeting room, lounge, and corridor with both simple and complex instructions.
As summarized in Table~\ref{tab:real_eval}, \modelname{} consistently achieves the best results, with significantly reduced navigation error (NE) and higher success rate (SR) across all three environments.

\paragraph{Qualitative real-world evaluation} As shown in Fig.~\ref{fig:realworld_vis}, \modelname{} exhibits coherent trajectories across meeting room, lounge, and corridor scenes.
It reliably reaches diverse targets such as chairs, sofas, and umbrellas, demonstrating robustness to clutter, narrow passages, and long-horizon paths in real-world settings.

\section{Test-Time Efficiency}
Test-time efficiency is critical for embodied models in practical robotics.
As illustrated in Fig.~\ref{fig:robot_setup}, the on-board Jetson Orin Nano faces strict resource limits, making large-scale inference prohibitively slow.
To address this, \modelname{} adopts a cloud-assisted design, where egocentric inputs are streamed to a remote server for reasoning, and only compact navigation commands are returned for execution.

We benchmark NaVid~\cite{zhang2024navid} and Uni-NaVid~\cite{zhang2024uninavid} on both on-board and server inference.
As shown in Table~\ref{tab:latency}, both baselines incur substantial delays on the Orin Nano, while server-side execution reduces latency to below 100\,ms.
\modelname{} achieves comparable efficiency with $\sim$95\,ms latency on the server, which is only slightly slower than NaVid and Uni-NaVid due to its dual-system reasoning overhead, yet this marginal gap does not affect real-time embodied navigation.

To ensure stable transmission, all experiments adopt a high-speed WiFi~6E (802.11ax, 6\,GHz band) network with 1.2\,Gbps peak bandwidth and $<$10\,ms access latency, covering the entire 200\,m$^2$ indoor test area.

\begin{table}[t]
\centering
\caption{\textbf{Average inference latency comparison.}
Comparison of average per-frame inference latency (ms) for NaVid, Uni-NaVid, and \modelname{} on Jetson Orin Nano and a remote server.
}
\resizebox{0.8\linewidth}{!}{
\begin{tabular}{lcc}
\toprule
Method & On-board (ms) & Server (ms) \\
\midrule
NaVid~\cite{zhang2024navid}        & 320 & 85 \\
Uni-NaVid~\cite{zhang2024uninavid} & 410 & 90 \\ \midrule
\textbf{\modelname{} (Ours)}       & --  & 95 \\
\bottomrule
\end{tabular}
}
\label{tab:latency}
\vspace{-0.4cm}
\end{table}

\section{Conclusion}
In this paper, we presented \modelname{}, an embodied foundation model designed to enhance both reasoning coherence and real-time navigation.
To overcome the instability of reasoning traces and the difficulty of balancing long-horizon semantics with real-time responsiveness, we introduced the large-scale \datasetname{} dataset for cold-start initialization, a GRPO-based reinforcement learning framework with format, understanding, and navigation rewards, and a Fast-in-Slow dual-system paradigm that decouples semantic reasoning from reactive control.
Comprehensive experiments on VLN, ObjectNav, embodied dialogue, planning, and reasoning benchmarks show that \modelname{} achieves consistent improvements in navigation success, trajectory fidelity, and reasoning coherence, while maintaining dialogue and planning performance on par with 3D-R1.
Moreover, real-world deployment on a Jetson Orin Nano-powered mobile robot further validates its robustness under limited edge resources, with cloud-assisted inference enabling real-time closed-loop control.

\clearpage
\section*{APPENDIX}

\section{Ablation Study}

\paragraph{Dual-system vs single-system} We compare the full dual-system \modelname{} with single-system variants that only retain either the slow semantic reasoning system or the fast reactive control system.
As shown in Table~\ref{tab:ablation_dualsystem}, both variants perform worse: the slow-only version struggles with real-time execution, while the fast-only version fails to maintain global semantic consistency.
The dual-system achieves the best trade-off, confirming the effectiveness of asynchronous coordination.

\begin{table}[h]
\centering
\caption{\textbf{Ablation on dual-system design.} Evaluation on R2R-CE Val-Unseen.}
\resizebox{0.8\linewidth}{!}{
\begin{tabular}{lcccc}
\toprule
Method & NE$\downarrow$ & OS$\uparrow$ & SR$\uparrow$ & SPL$\uparrow$ \\
\midrule
Slow-only          & 5.12 & 68.3 & 61.2 & 54.3 \\
Fast-only          & 5.47 & 65.1 & 58.7 & 50.1 \\ \midrule
\textbf{Dual-system (Ours)} & \textbf{3.86} & \textbf{74.1} & \textbf{72.5} & \textbf{68.8} \\
\bottomrule
\end{tabular}
}
\label{tab:ablation_dualsystem}
\end{table}

\paragraph{Reward decomposition} To verify the effectiveness of the proposed reward design, we ablate the three rewards in RL training: format reward $R_{\text{Format}}$, understanding reward $R_{\text{Understanding}}$, and navigation reward $R_{\text{Navigation}}$.
Table~\ref{tab:ablation_rewards} shows that removing any reward leads to performance degradation.
Without $R_{\text{Format}}$, the model often generates unstructured outputs. Removing $R_{\text{Understanding}}$ reduces semantic grounding, while dropping $R_{\text{Navigation}}$ severely hurts trajectory fidelity.
The full combination achieves the best results, demonstrating that the three rewards are complementary.

\begin{table}[h]
\centering
\caption{\textbf{Reward decomposition on HM3D-OVON Val-Unseen.} \cmark denotes inclusion of the reward.}
\resizebox{0.8\linewidth}{!}{
\begin{tabular}{ccccc}
\toprule
$R_{\text{Format}}$ & $R_{\text{Understanding}}$ & $R_{\text{Navigation}}$ & SR$\uparrow$ & SPL$\uparrow$ \\
\midrule
\xmark & \xmark & \xmark & 34.5 & 15.2 \\
\cmark & \xmark & \xmark & 36.1 & 16.0 \\
\xmark & \cmark & \xmark & 37.0 & 16.7 \\
\xmark & \xmark & \cmark & 36.5 & 15.9 \\
\midrule
\cmark & \cmark & \xmark & 39.4 & 18.2 \\
\cmark & \xmark & \cmark & 38.7 & 17.5 \\
\xmark & \cmark & \cmark & 39.9 & 18.7 \\
\midrule
\cmark & \cmark & \cmark & \textbf{42.2} & \textbf{20.1} \\
\bottomrule
\end{tabular}
}
\label{tab:ablation_rewards}
\end{table}

\paragraph{Hyper-parameters} We further study the sensitivity of \modelname{} to hyper-parameters on the RxR-CE Val-Unseen split. Tables~\ref{tab:ablation_beta} report the ablation result.
For the KL penalty $\beta$, too small values cause divergence from the reference policy, while too large values limit exploration.
The best trade-off is achieved at $\beta=0.02$.

\begin{table}[h]
\centering
\caption{\textbf{Ablation on KL penalty $\beta$.} Evaluation on RxR-CE Val-Unseen.}
\resizebox{0.75\linewidth}{!}{
\begin{tabular}{lcccc}
\toprule
$\beta$ & NE$\downarrow$ & SR$\uparrow$ & SPL$\uparrow$ & nDTW$\uparrow$ \\
\midrule
0.005 & 5.12 & 64.3 & 59.5 & 70.1 \\
0.01  & 4.56 & 66.3 & 60.7 & 72.9 \\ \midrule
\textbf{0.02 (Ours)} & \textbf{3.98} & \textbf{71.3} & \textbf{66.3} & \textbf{79.4} \\ \midrule
0.03  & 4.23 & 68.9 & 64.2 & 75.1 \\
0.05 & 4.36 & 67.8 & 62.4 & 74.2 \\
\bottomrule
\end{tabular}
}
\label{tab:ablation_beta}
\end{table}

\section{Limitation and Future Work}
Although \modelname{} achieves strong results, it still has several limitations.
The \datasetname{} dataset, though large, is mainly synthesized from existing benchmarks and does not fully capture real-world complexity.
Our model also relies on RGB-D and language inputs, without integrating richer modalities such as audio or tactile signals.
Moreover, current deployment depends on cloud inference, limiting real-time scalability on edge devices.
Future work will explore expanding data coverage, incorporating multimodal perception, improving efficiency for on-board deployment, and extending to longer-horizon and more diverse embodied tasks.

\section{Visualization}
\begin{figure*}[ht]
    \centering
    \includegraphics[width=0.8\linewidth]{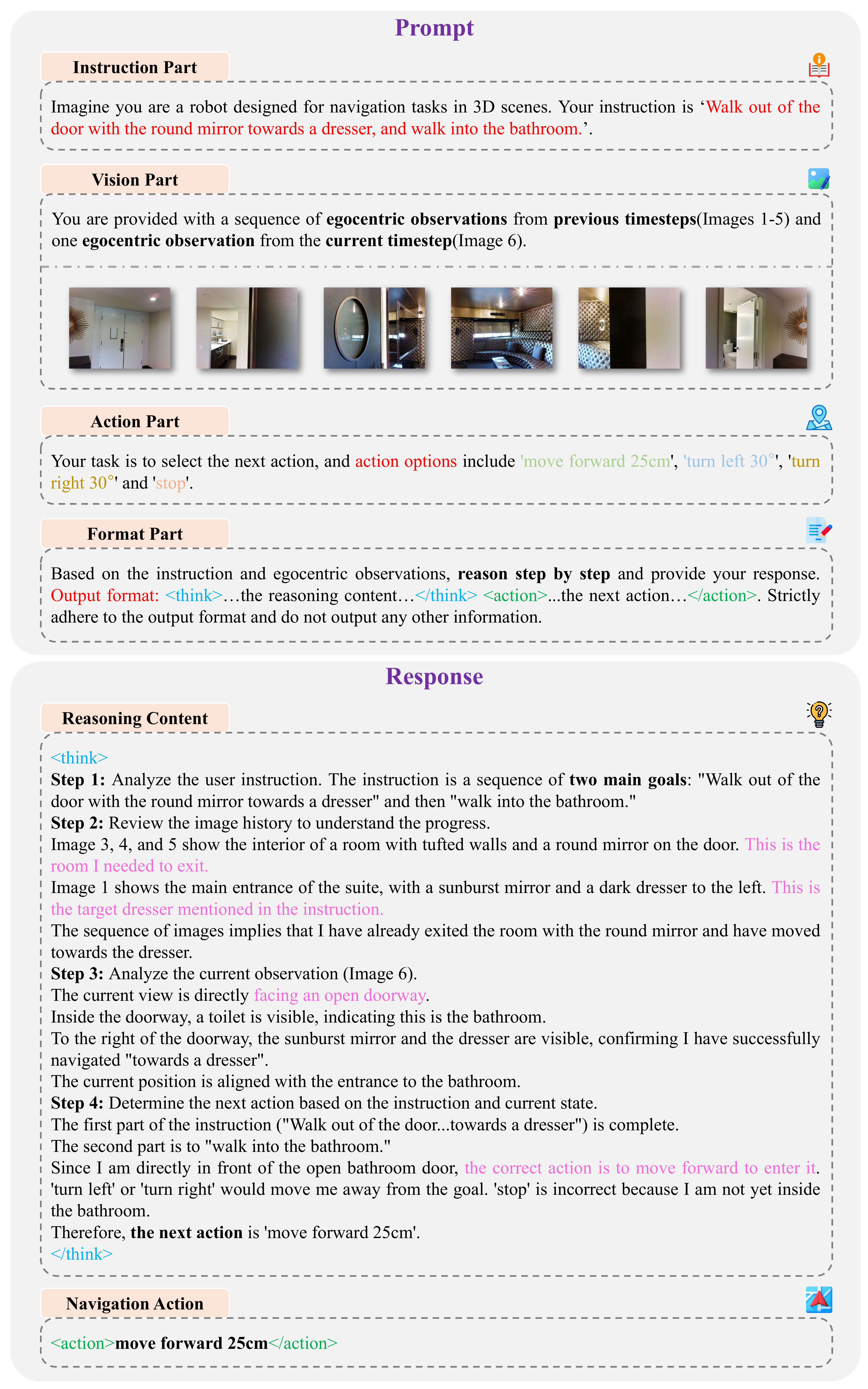}
    \caption{\textbf{\datasetname{} CoT data example.}
    }
    \label{fig:cot_example}
\end{figure*}

\paragraph{CoT example}
To better illustrate how structured reasoning traces are generated, Fig.~\ref{fig:cot_example} visualizes a representative \datasetname{} example. 
It shows how natural language instructions, egocentric observations, and candidate actions are transformed into step-by-step reasoning content and navigation decisions, highlighting the role of CoT supervision in stabilizing model training.

\begin{figure*}[ht]
    \centering
    \includegraphics[width=0.75\linewidth]{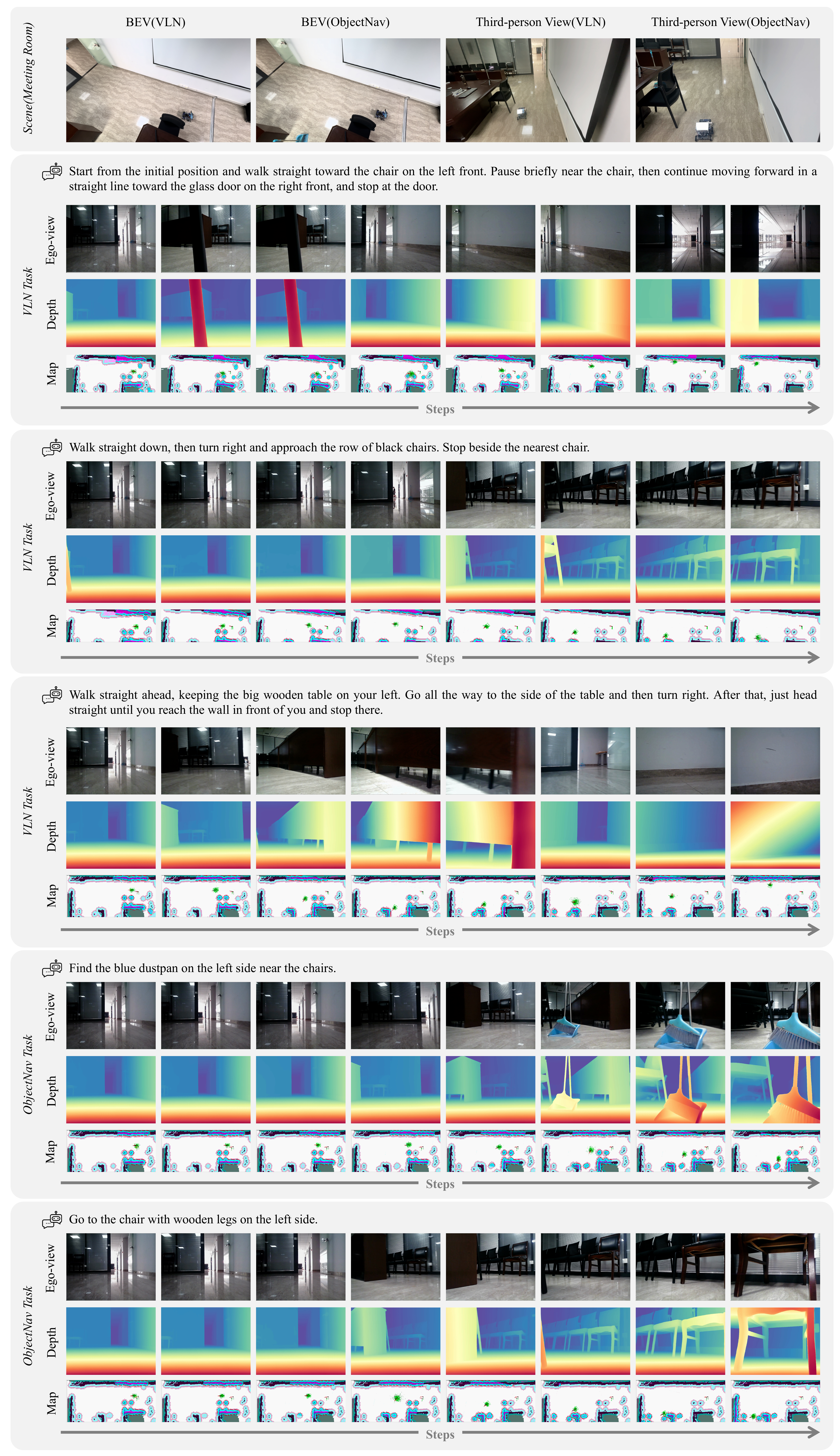}
    \caption{\textbf{Real-world qualitative results of \modelname{} on VLN and ObjectNav tasks in meeting room.}
    }
    \label{fig:realworld_vis_meeting}
\end{figure*}

\begin{figure*}[ht]
    \centering
    \includegraphics[width=0.75\linewidth]{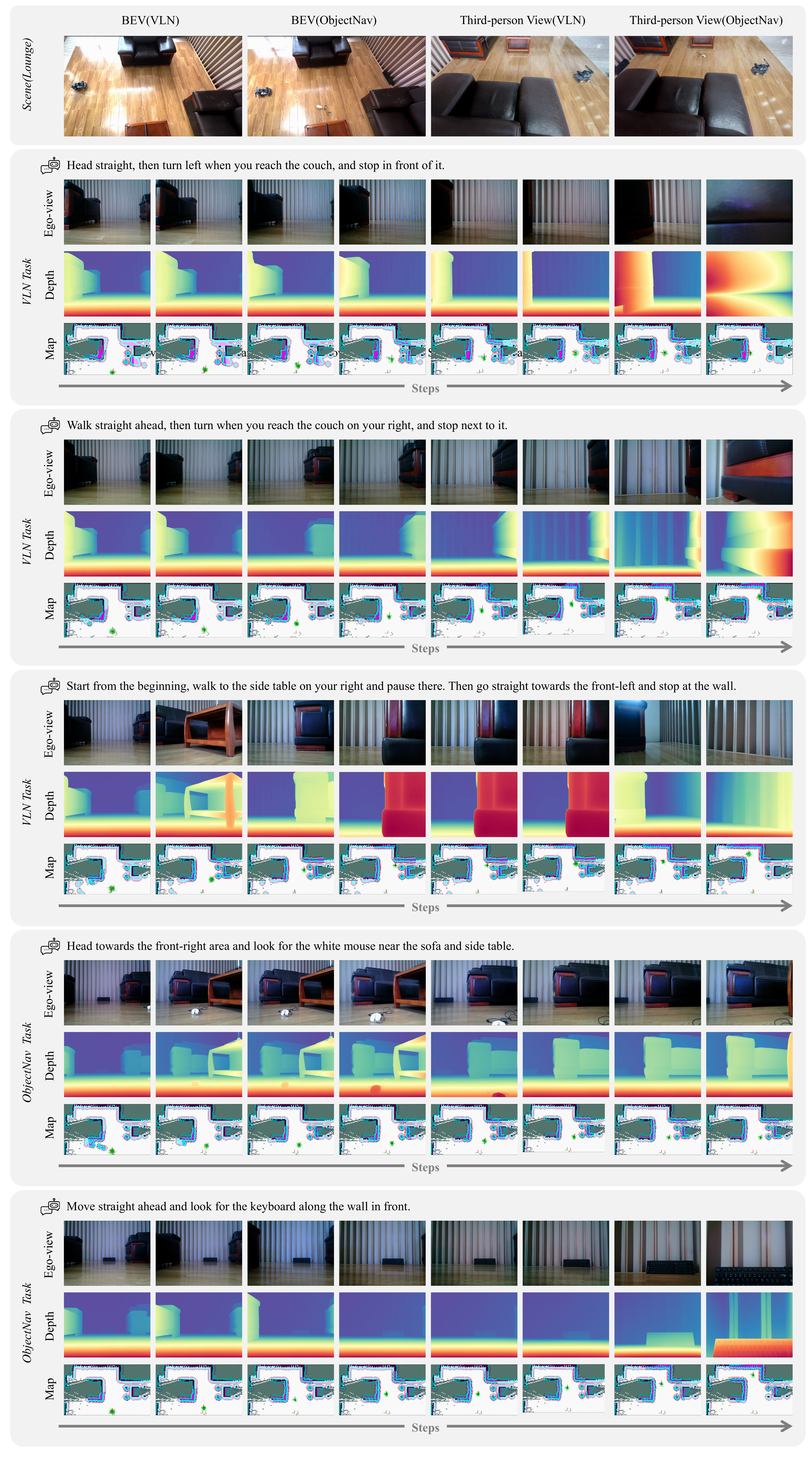}
    \caption{\textbf{Real-world qualitative results of \modelname{} on VLN and ObjectNav tasks in lounge.}
    }
    \label{fig:realworld_vis_lounge}
\end{figure*}

\begin{figure*}[ht]
    \centering
    \includegraphics[width=0.75\linewidth]{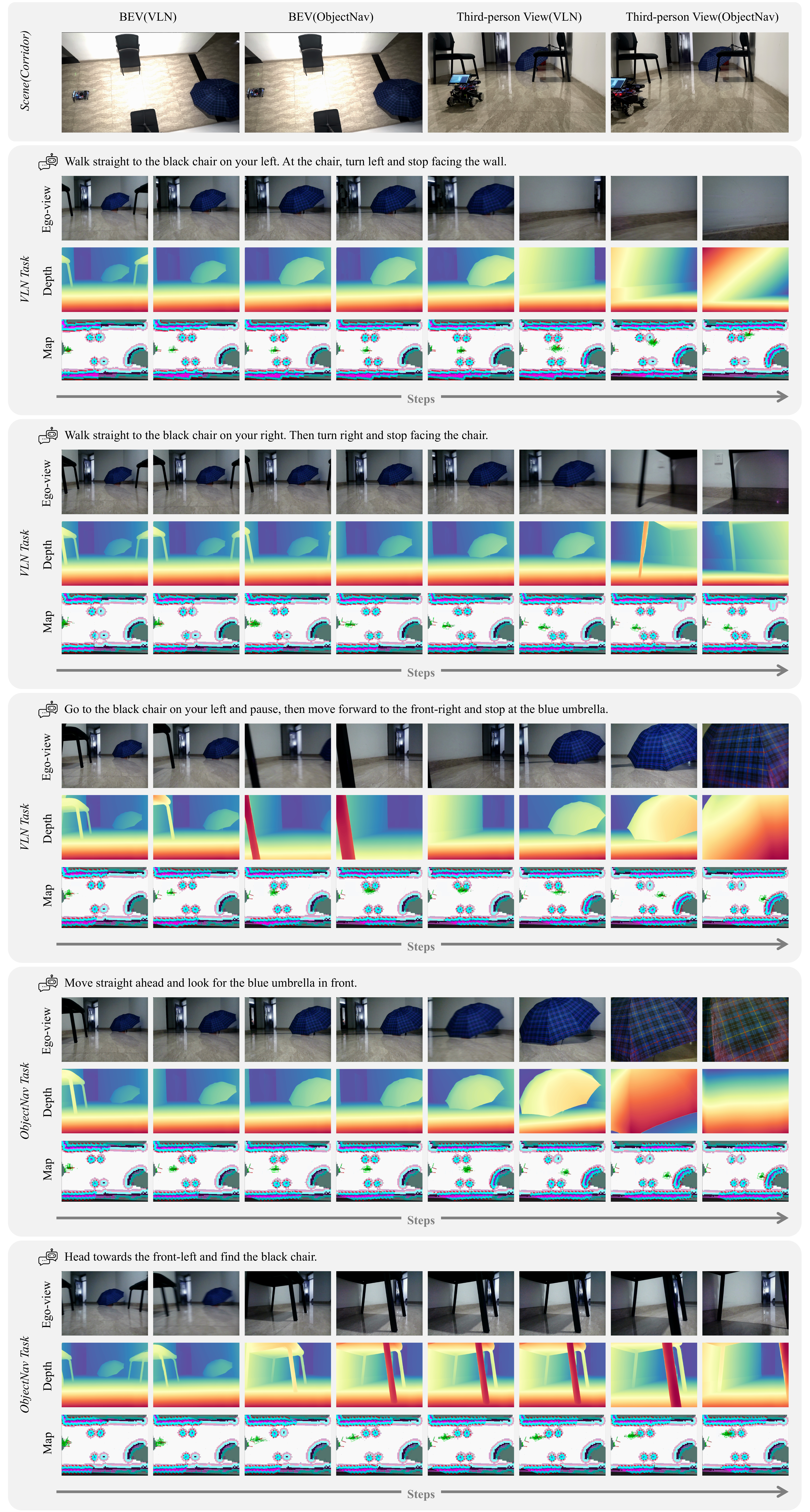}
    \caption{\textbf{Real-world qualitative results of \modelname{} on VLN and ObjectNav tasks in corridor.}
    }
    \label{fig:realworld_vis_corridor}
\end{figure*}

\paragraph{Results visualization}
To further illustrate the real-world performance of \modelname{}, we provide qualitative visualizations across multiple embodied tasks and diverse indoor environments.
As shown in Fig.~\ref{fig:realworld_vis_meeting}–\ref{fig:realworld_vis_corridor}, we first evaluate navigation-oriented behaviors in three distinct scenes. Each example depicts the natural language instruction, egocentric RGB observations, depth maps, and LiDAR-based top-down maps, complemented with third-person views and BEV trajectories for reference.

\begin{figure*}[ht]
    \centering
    \includegraphics[width=0.9\linewidth]{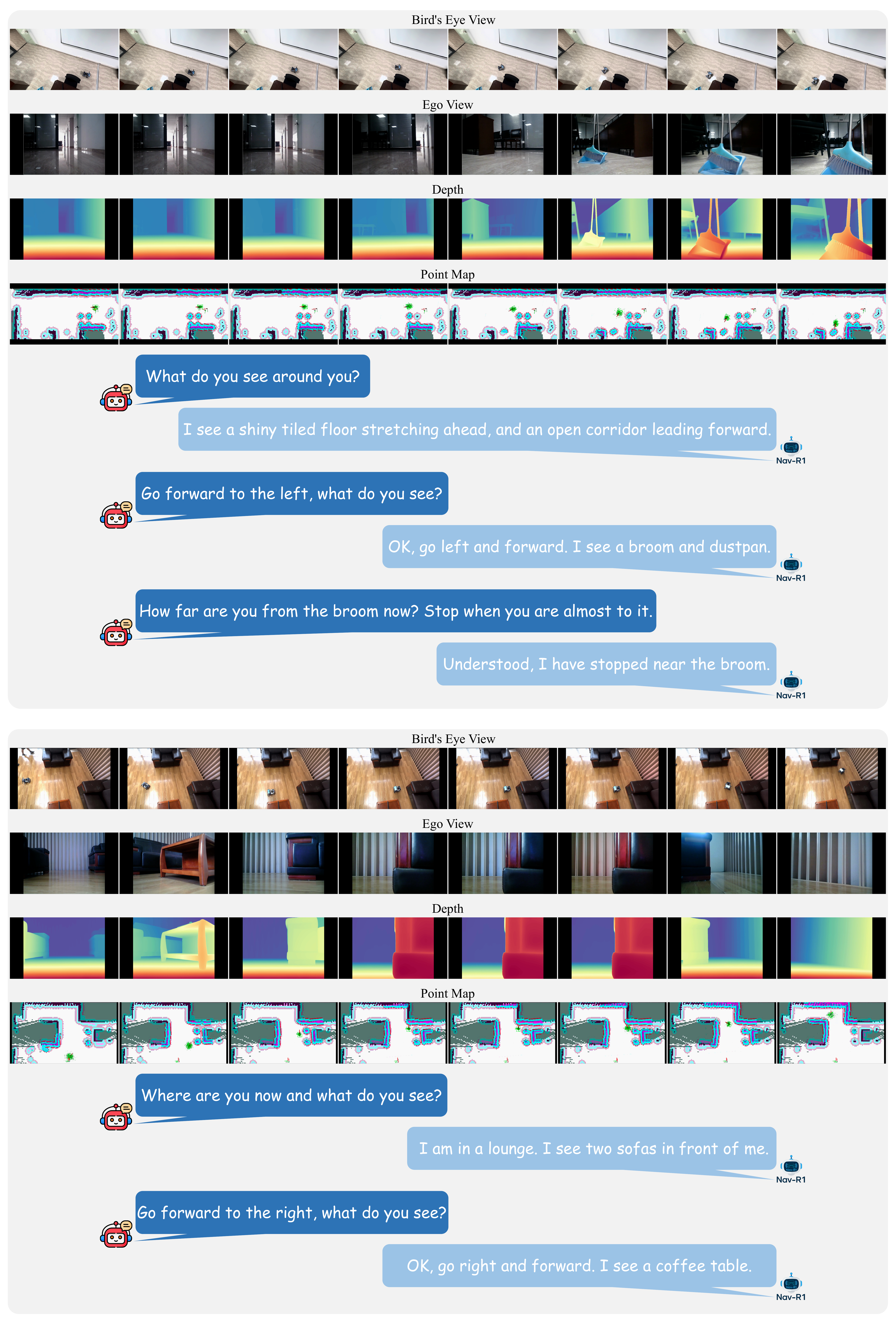}
    \caption{\textbf{Real-world qualitative results of \modelname{} on embodied dialogue task.}
    }
    \label{fig:realworld_vis_dialogue}
\end{figure*}

\begin{figure*}[ht]
    \centering
    \includegraphics[width=0.9\linewidth]{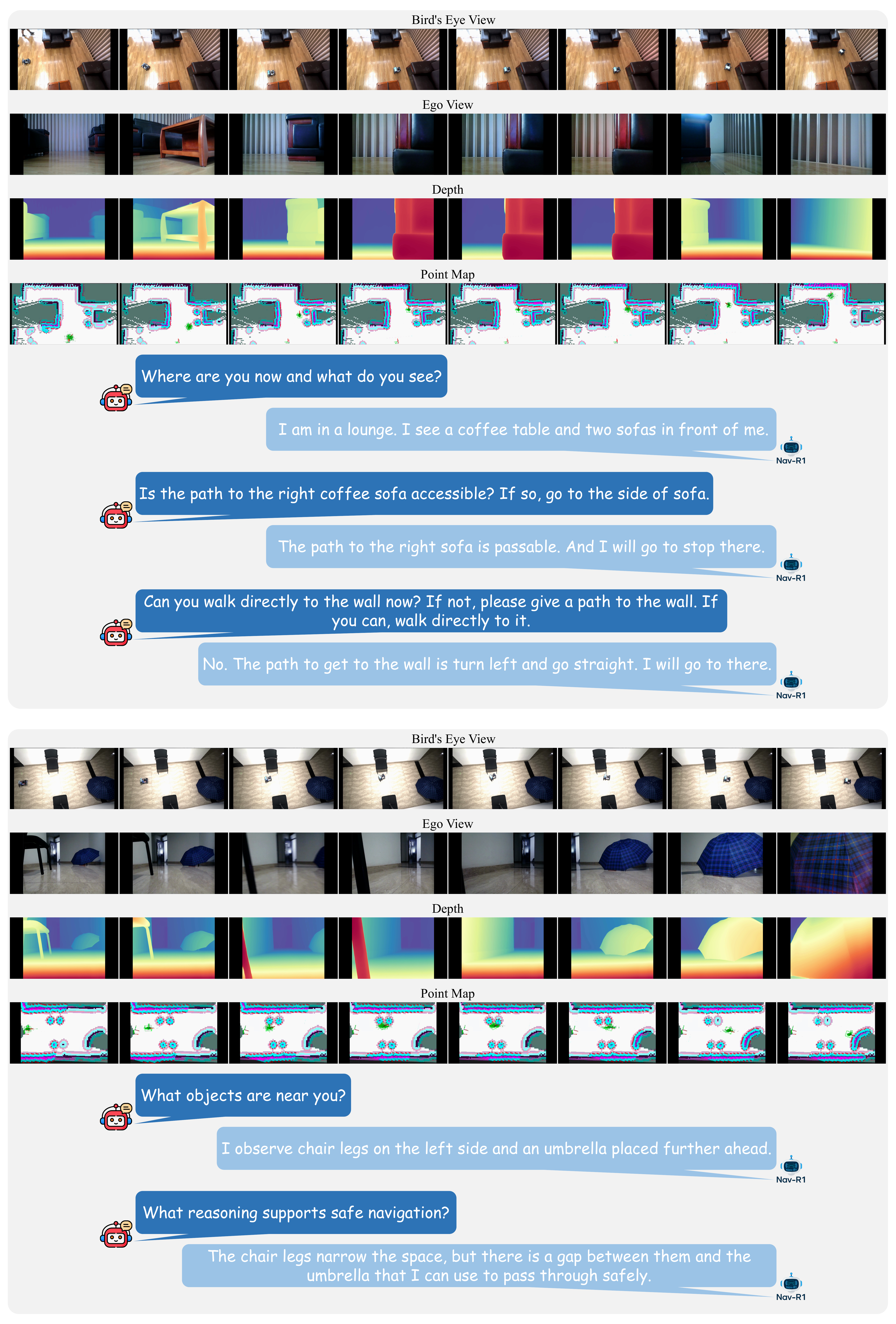}
    \caption{\textbf{Real-world qualitative results of \modelname{} on embodied reasoning task.}
    }
    \label{fig:realworld_vis_reasoning}
\end{figure*}

\begin{figure*}[ht]
    \centering
    \includegraphics[width=0.9\linewidth]{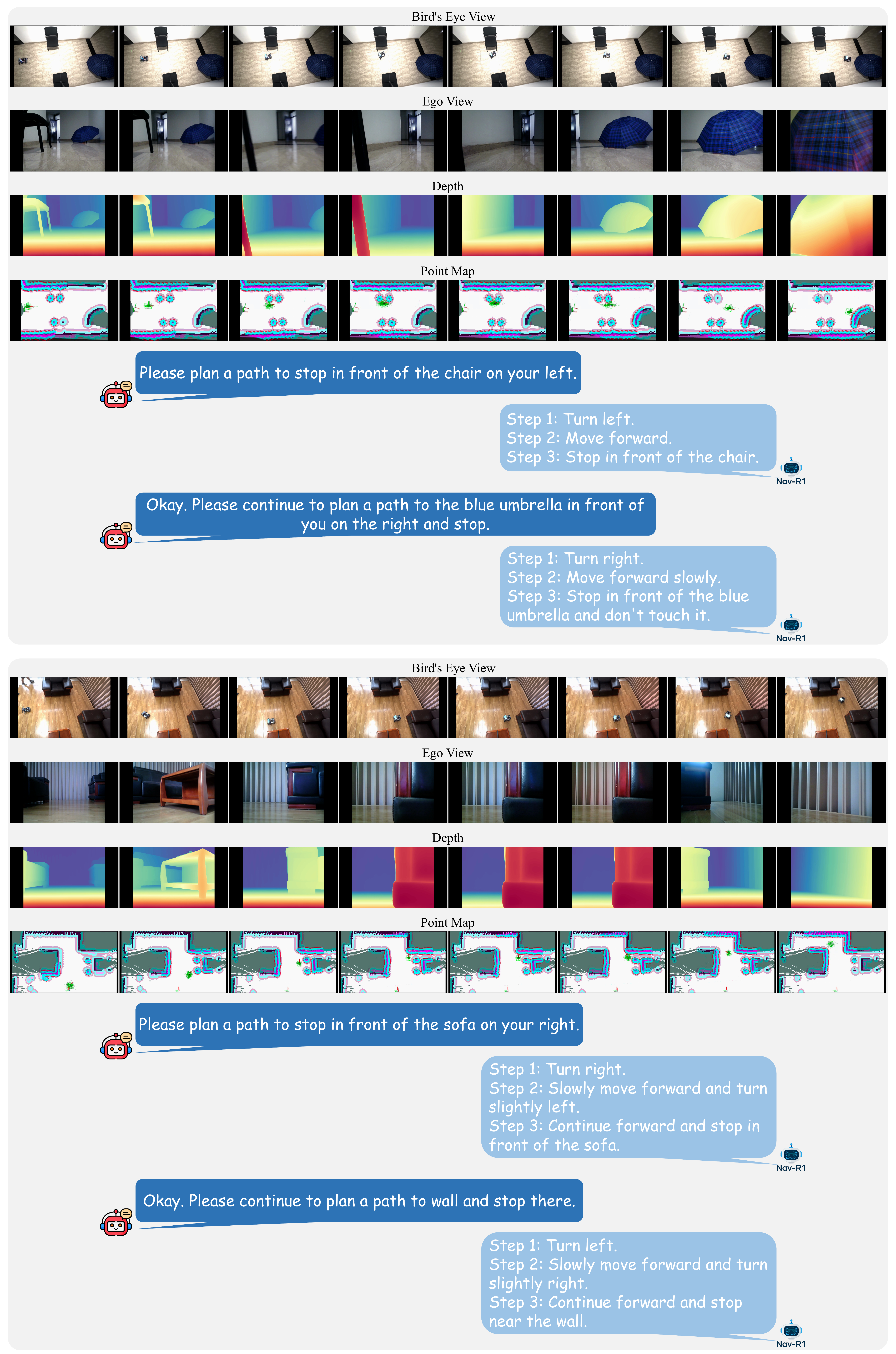}
    \caption{\textbf{Real-world qualitative results of \modelname{} on embodied planning task.}
    }
    \label{fig:realworld_vis_planning}
\end{figure*}

Beyond navigation, we further demonstrate the versatility of \modelname{} in dialogue, reasoning, and planning tasks.
Fig.~\ref{fig:realworld_vis_dialogue}–\ref{fig:realworld_vis_planning} present real-world qualitative results on embodied dialogue, reasoning, and planning, respectively.
These visualizations highlight that \modelname{} not only executes goal-directed trajectories robustly, but also maintains coherent interaction, safe reasoning, and multi-step planning capabilities under complex real-world layouts.

\begin{figure*}[ht]
    \centering
    \includegraphics[width=\linewidth]{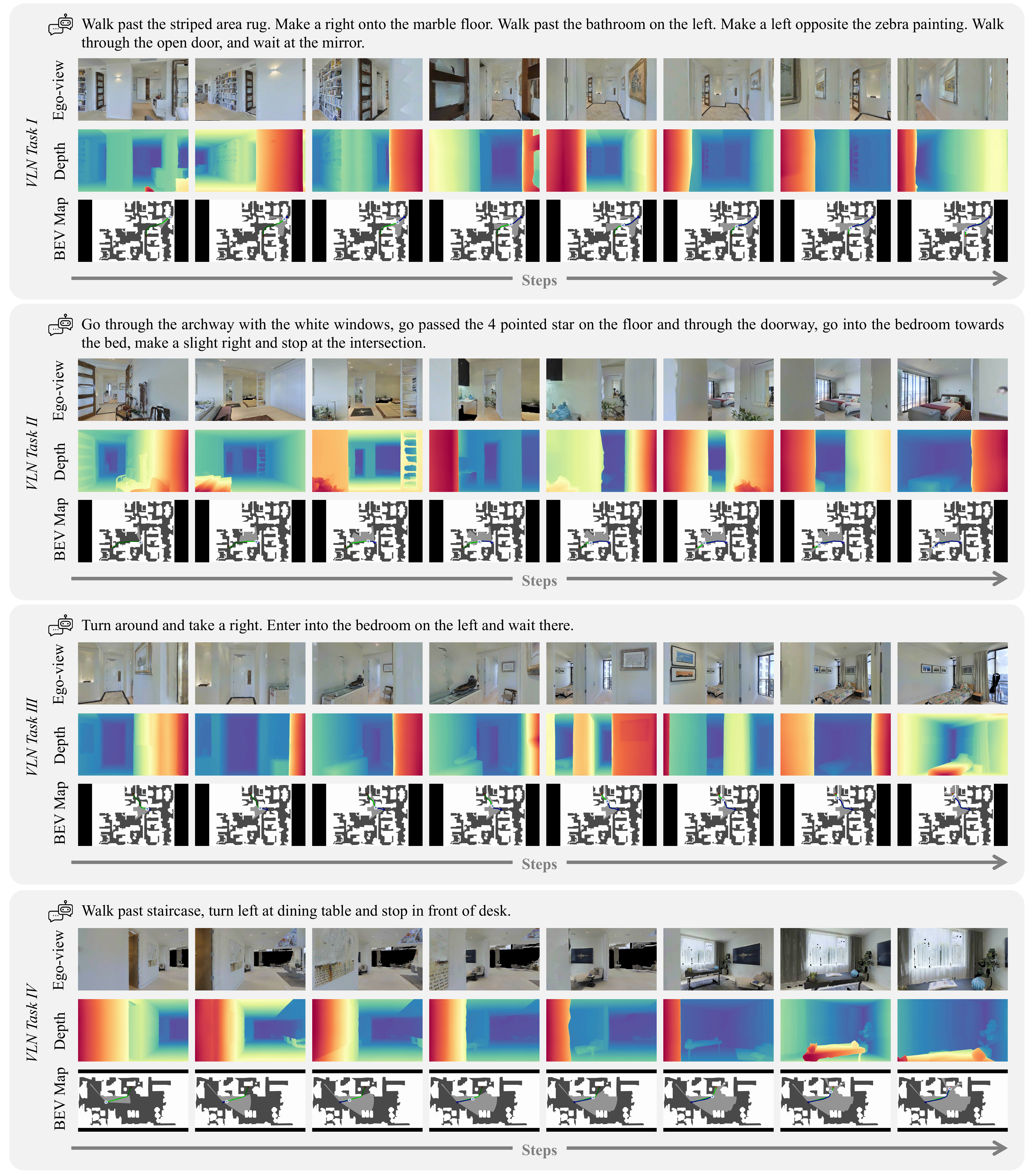}
    \caption{\textbf{Visual results of VLN on VLN-CE R2R.}
    }
    \label{fig:simulator_vln_1}
\end{figure*}

\begin{figure*}[ht]
    \centering
    \includegraphics[width=\linewidth]{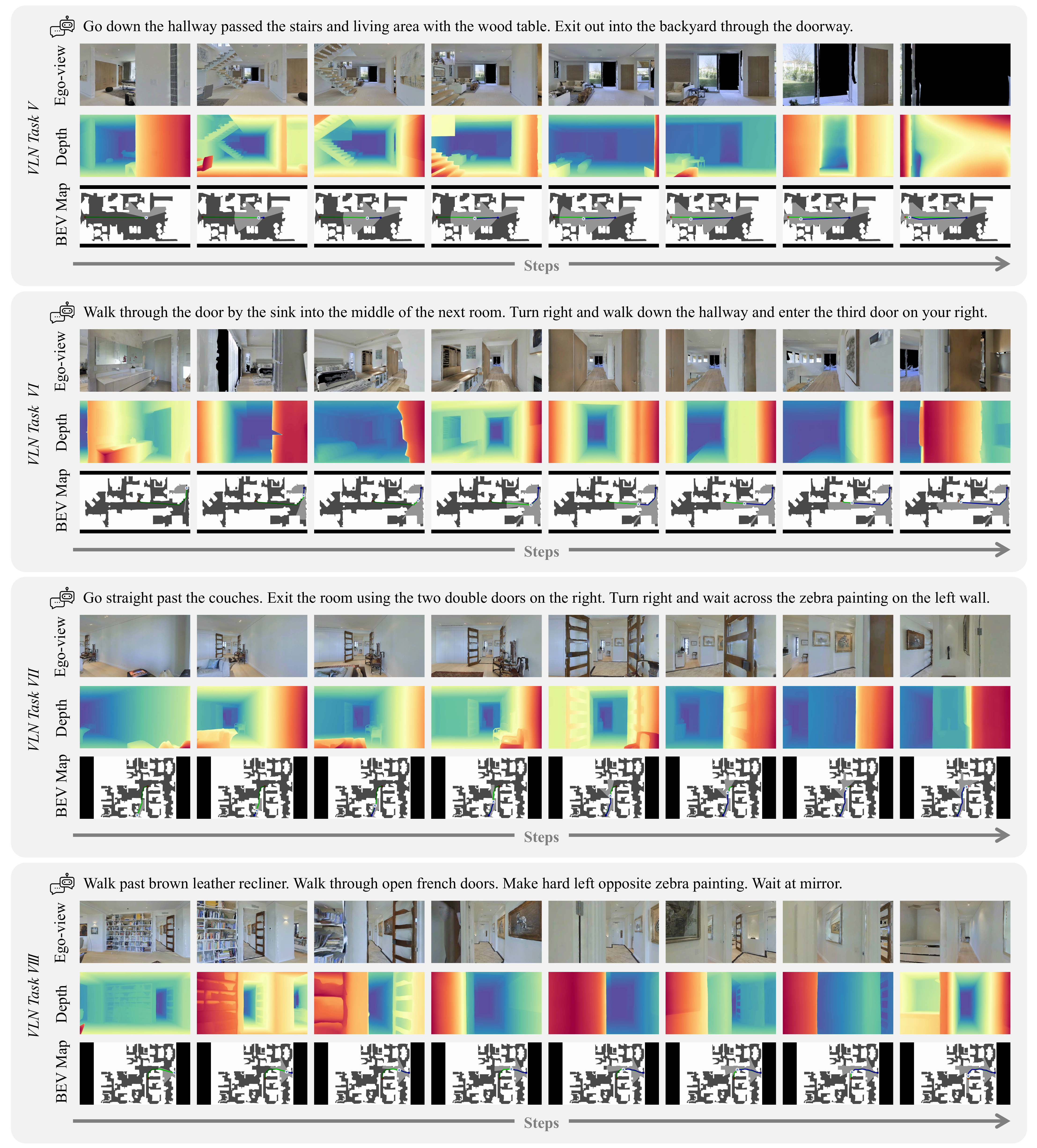}
    \caption{\textbf{Visual results of VLN on VLN-CE R2R.}
    }
    \label{fig:simulator_vln_2}
\end{figure*}

\begin{figure*}[ht]
    \centering
    \includegraphics[width=\linewidth]{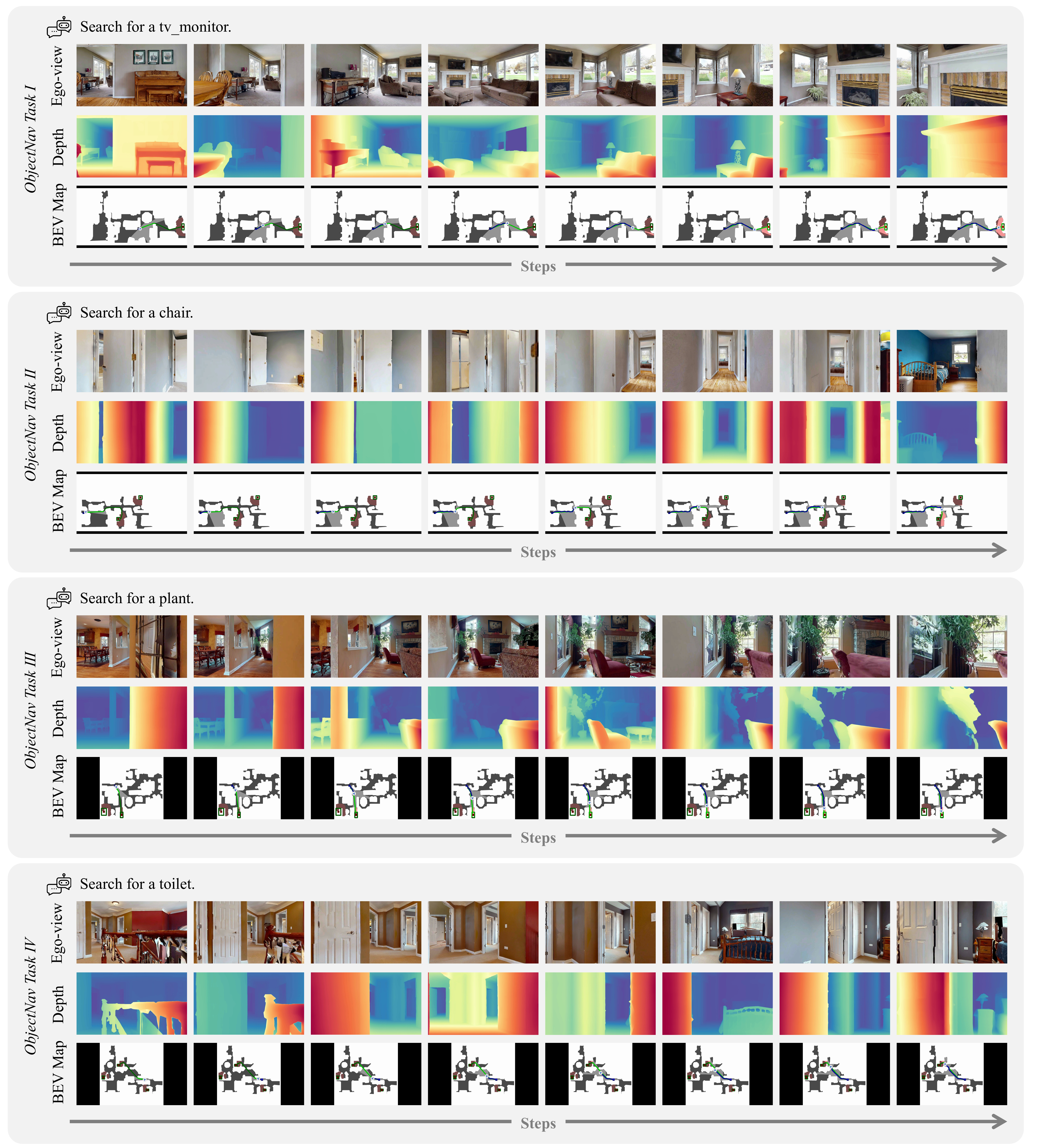}
    \caption{\textbf{Visual results of ObjectNav on HM3D.}
    }
    \label{fig:simulator_objnav}
\end{figure*}

In addition to real-world deployment, we also provide benchmark visualizations on simulation datasets.
Fig.~\ref{fig:simulator_vln_1} and Fig.~\ref{fig:simulator_vln_2} highlight the VLN-CE R2R benchmark, where \modelname{} successfully grounds long-horizon instructions into coherent navigation trajectories.
Similarly, Fig.~\ref{fig:simulator_objnav} presents results on HM3D ObjectNav, showing that the agent can robustly explore large-scale 3D layouts and accurately localize target objects.
These benchmark results further confirm that the proposed model generalizes across navigation paradigms.

\end{document}